\documentclass[runningheads]{llncs}
\usepackage{multirow}
 
\usepackage[year=2026,ID=14135]{eccv}

\usepackage{eccvabbrv}

\usepackage{graphicx}
\usepackage{booktabs}

\usepackage[accsupp]{axessibility}  

\usepackage{hyperref}

\usepackage{orcidlink}

\begin{document}

\title{GeoCFM: Positive-Only Conditional Flow Matching for Mineral Occurrence Sampling} 

\titlerunning{GeoCFM: Positive-Only CFM for Mineral Occurrence Sampling}

\author{Moshe Eliasof\inst{1}\thanks{Corresponding author.} \and
Eldad Haber\inst{2}}

\authorrunning{M.~Eliasof and E.~Haber}

\institute{Faculty of Computer and Information Science, Ben-Gurion University of the Negev, Israel\\
\email{eliasof@bgu.ac.il} \and
Department of Earth, Ocean and Atmospheric Sciences,\\ University of British Columbia, Canada\\
\email{eldadhaber@gmail.com}}

\maketitle

\begin{abstract}
Critical mineral discovery is a positive-only problem: deposits are observed as sparse locations, while unlabeled regions are not reliable negatives, and similar geophysical signatures can arise from different subsurface states. We therefore model mineral targeting as learning a conditional spatial distribution over occurrence locations, \(\pi(p\mid d)\), given geo-images \(d\), rather than predicting a deterministic per-pixel score map. We introduce {GeoCFM}, a conditional flow-matching model that generates mineral occurrence point sets conditioned on multi-channel geo-images; GeoCFM learns a point-wise transport field in \(\mathbb{R}^2\), using UNet features with point-conditioned velocity prediction to bridge dense rasters and sparse supervision without pseudo-negatives. On a synthetic magnetics--geochemistry benchmark with latent activation and on USGS Earth MRI data with a spatially disjoint tile split, GeoCFM improves geometric agreement with observed occurrences over score-map and non-conditional baselines, while representing epistemic uncertainty through conditional sampling.
\end{abstract}

\section{Introduction}
\label{sec:introduction}
Mineral exploration is decision-making under extreme uncertainty. Economically viable mineralization is spatially rare, geologically complex, and only indirectly observed through noisy and heterogeneous measurements. In practice, exploration workflows fuse geological, geophysical, and geochemical data layers into mineral discovery maps, also known as mineral prospectivity, that rank targets \cite{BonhamCarter1994,Carranza2009,Zuo2020}. Despite advances in sensing, modeling, and machine learning, the global discovery rate of significant mineral deposits has declined over the past decade from a late-2000s peak \cite{Schodde2025}, while drilling remains expensive, irreversible, environmentally damaging and high risk \cite{SingerKouda1999}.

\begin{figure}[t]
    \centering
    \includegraphics[width=0.3\linewidth]{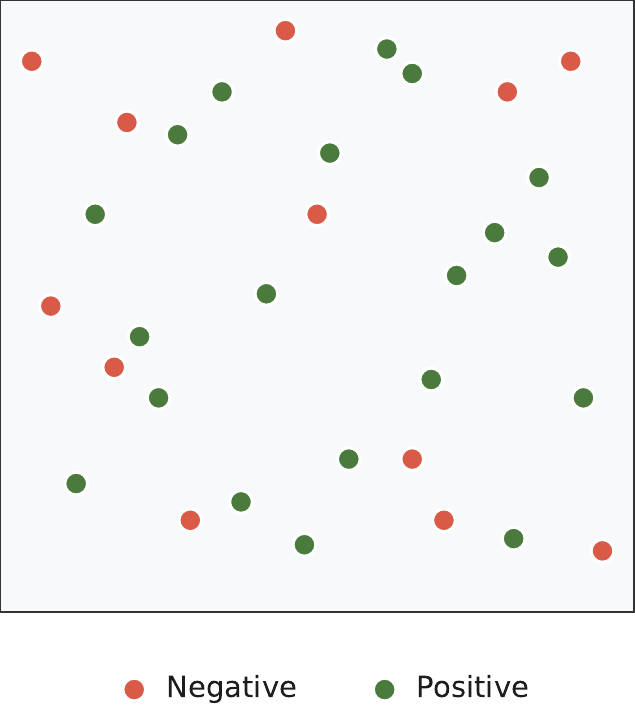}\hfill
    \includegraphics[width=0.3\linewidth]{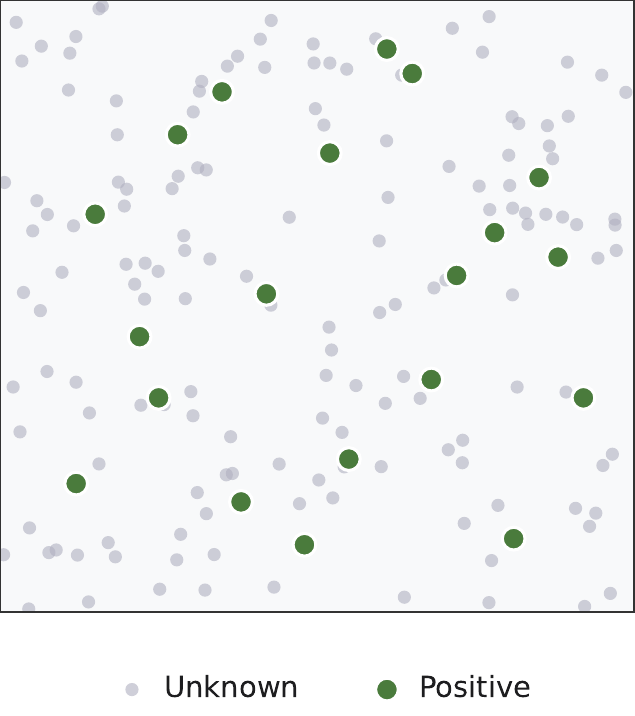}\hfill
    \includegraphics[width=0.3\linewidth]{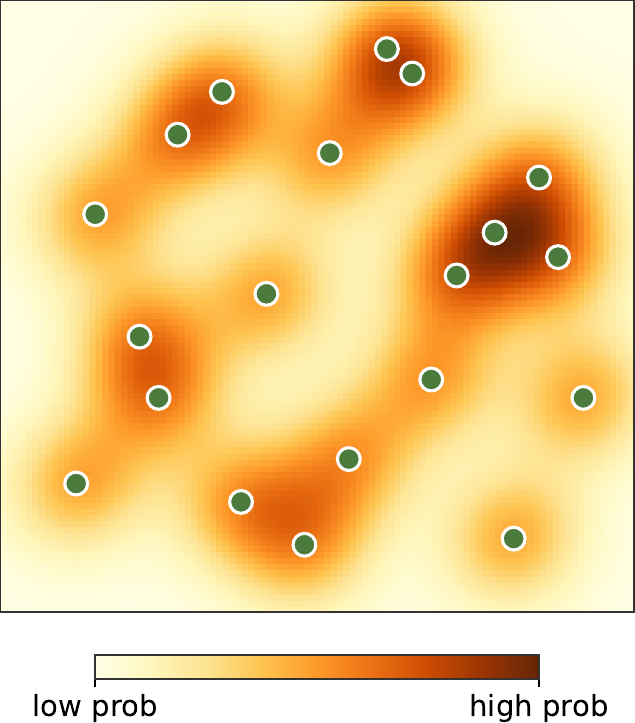}
    
    \caption{\textbf{Mineral prospectivity as a Positive-Only learning problem.} Left: Standard supervised learning relies on both known positives and confirmed negatives to learn a decision boundary. Middle: The reality of mineral exploration is positive-only; we observe sparse positive deposits, but the remaining area is unknown/unlabeled (absence of a mine is not evidence of absence). Right: Our GeoCFM approach models a conditional density over mineral occurrences given geo-images, directly yielding a probabilistic mineral discovery map from positive samples alone.}
    \label{fig:pu_schematic}
\end{figure}

A key difficulty is that mineral exploration is \emph{not} a standard supervised learning problem. We typically observe only a limited set of labeled positives (known mines and mineral anomaly occurrences), but we do not observe trustworthy negatives: an unlabeled location may be untested, undiscovered, or simply uneconomic under current conditions. The supervision is therefore inherently \emph{positive-only} (PO), and in practice \emph{positive and unlabeled} (PU) when the remaining spatial search space is treated as unlabeled data \cite{ElkanN08,KiryoNPS17}. A common workaround in data-driven prospectivity mapping is to sample pseudo-negatives from unlabeled regions, but many such samples are not true negatives, which can systematically bias the learned decision boundary \cite{ZuoWang2020}. This issue is compounded by historical exploration bias: available labels and supporting datasets are concentrated in mature, historically targeted areas, so treating unlabeled regions as negatives amplifies uneven sampling of the search space \cite{HronskyKreuzer2019}. We summarize this mismatch between the \emph{true} supervision and the \emph{assumed} supervision in \Cref{fig:pu_schematic}.

Beyond missing negatives, mineral exploration is dominated by epistemic uncertainty: the same surface and near-surface measurements can be consistent with {\em multiple plausible subsurface states}. 

\begin{figure}[t]
    \centering
\includegraphics[width=0.82\linewidth]{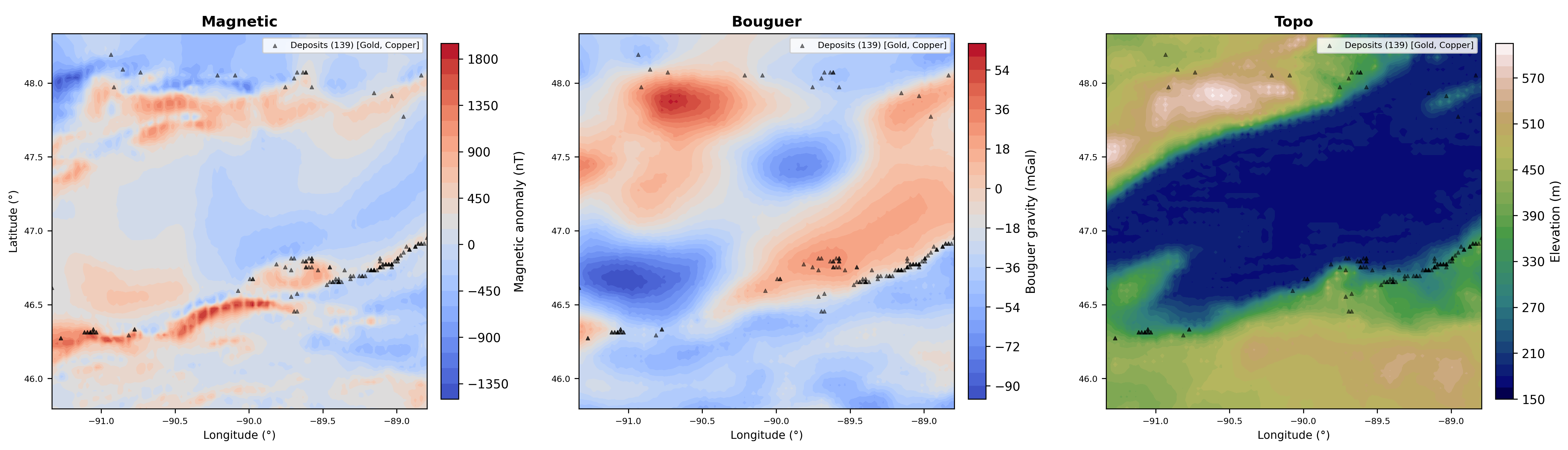}
    \caption{Magnetics (left), gravity (middle), and topography (right) together with mineral (gold or copper) occurrences, i.e., deposits, in a geo-image over North America.}
    \label{fig:minerals}
\end{figure}

Formally, let \(d\) denote the observable geoscience measurements (geo-images), e.g., magnetic and gravity grids, geochemical assays, structural geological images, and other sensing modalities, as illustrated in \Cref{fig:minerals}. These measurements provide only indirect and partial constraints on the subsurface. We  treat the rest of the geological states as a latent variable \(g\in\mathcal{G}\), where \(\mathcal{G}\) is a high-dimensional geological manifold encoding the lithological structure, fluid pathways, the history of alteration and structural evolution of the Earth that are unobservable. Under this view, mineralization is modeled as the outcome of a mapping
\begin{equation}
\label{eq:mineralization}
p = f(d, g), \qquad g \in \mathcal{G},
\end{equation}
where $p$ denotes the position of a mineral occurrence. Hence, the inverse mapping \(d \mapsto p\) is inherently one-to-many. This non-identifiability induces epistemic uncertainty, and motivates probabilistic predictions rather than a deterministic label map.

Mineral occurrences can be viewed as sparse \emph{keypoints} over geo-image inputs. The central question is not whether an anomaly ``contains mineralization,'' but how to assign a \emph{calibrated probability of mineralization} given the evidence and its uncertainties. Unlike dense prediction (e.g., semantic segmentation), where labels define an (approximately) deterministic pixel-to-class mapping, mineral discovery is one-to-many: similar sensor patterns may correspond to different latent subsurface states, and only a subset yields economic mineralization. The desired output is therefore \emph{probabilistic}, enabling sampling over plausible subsurface outcomes. 
From a learning perspective, this shifts the objective from discriminative prediction to conditional density sampling. Rather than learning a classifier with \emph{pseudo} negatives, we model a distribution of mineralization outcomes and sample multiple hypotheses consistent with the same evidence. Concretely, we represent deposits by their spatial coordinates \(p\) and learn a flow that samples \(\pi(p \mid d)\) from positive occurrences alone. Sampling from \(\pi(p \mid d)\) exposes epistemic uncertainty, supports uncertainty-aware target ranking, and avoids collapsing ambiguous signals into a single prediction.

In this work, we cast mineral discovery as probabilistic inference under positive-only supervision and materialize it with \textbf{GeoCFM}, a conditional flow-matching model that generates mineral occurrence point sets conditioned on geo-images. GeoCFM parameterizes the conditional velocity field with a shared MLP, guided by local features extracted from the geo-image by a UNet. Starting from a 2D Gaussian point cloud, the learned flow transports points into spatially resolved occurrence locations, without requiring any negative samples.

\noindent\textbf{Contributions.} Our contributions are:
\begin{enumerate}
\item We introduce GeoCFM -- a positive-only, image-conditioned point sampler for mineral discovery. Our key contribution is not positive-only learning per se, but casting prospectivity as sampling from a conditional spatial density \(\pi(p\mid d)\) that avoids explicit negatives while remaining a sampleable, multi-modal predictive object, instantiated via conditional flow matching with point-conditioned UNet features.
\item We propose an image-to-points architecture that conditions a point-wise velocity field on geo-images via local UNet features sampled at point locations, bridging dense rasters and sparse occurrence supervision without pseudo-negatives.
\item We provide a practical training and evaluation protocol for spatial geoscience data, including a multi-metric, repeated-sampling evaluation (Chamfer distance, Sinkhorn optimal transport, F@5, KDE negative log-likelihood, and top-5\% hit rate) and a spatially disjoint remote-sensing split.
\item We show on a controlled synthetic benchmark and on real USGS Earth MRI data that sampling-based mineral discovery captures multi-modality and improves over a broad baseline suite spanning score-map, boosting, one-class, retrieval, and non-conditional methods.
\end{enumerate}


\section{Related Work}
We briefly review prior work on (i) Mineral discovery/mineral prospectivity mapping, 
(ii) positive-only and PU learning, and (iii) conditional generative modeling via transport, which motivates our flow-matching formulation.

\paragraph{Mineral Discovery/Prospectivity mapping.}
Prospectivity mapping traditionally fuses heterogeneous evidence from geology, geophysics, and geochemistry into a spatial {favorability} or {prospectivity} score. Classic knowledge-driven workflows use expert rules and evidential weighting, while data-driven variants fit statistical or discriminative models to known occurrences, ranging from logistic and Bayesian-style formulations to random forests and neural networks \cite{BonhamCarter1994,Carranza2009,Cracknell01122015,GranekHaber2015SDM,GranekHaber2016GeoscienceBC,McMillanTLE}. Regardless of modeling choice, most pipelines output a single deterministic score map that is subsequently thresholded or ranked to prioritize follow-up. Since confirmed negatives are rarely available at scale, these methods often introduce pseudo-negatives by sampling from unlabeled regions. When supervision is in fact positive-only, pseudo-negatives can include undiscovered or untested positives and can be spatially biased by historical sampling, which distorts the learned decision rule and undermines calibration \cite{ZuoWang2020}. A recurring response is to avoid explicit negatives altogether: one-class formulations such as one-class SVMs estimate the support of the positive distribution \cite{ChenWuZhao2019OCSVM}, while more recent work explores self-supervised and explainability-oriented pipelines that target scalability and uncertainty in exploration \cite{DarunaEtAl2024ScalableMineralExploration}. These approaches still produce a per-pixel favorability score; in contrast, we model occurrences as samples from a conditional spatial density, which yields a sampleable, multi-modal predictive object rather than a single deterministic map. We compare against representative score-map, boosting, one-class, and retrieval baselines in \Cref{sec:experiments}.

\paragraph{Positive-only and PU learning.}
Learning from positive and unlabeled data (PU learning) is a well-studied weak-supervision setting in which the unlabeled set is a mixture of positives and negatives \cite{ElkanN08,BekkerDavis2020Survey,KiryoNPS17}. Core tools include risk estimators that avoid treating unlabeled samples as negatives, together with class-prior (mixture proportion) estimation to convert positive-conditional models into calibrated posteriors \cite{duPlessisSugiyama2014,duPlessisSugiyama2017,RamaswamyScottTewari2016,IyerNathSarawagi2014}. Much of the PU literature is developed under assumptions such as positives being labeled completely at random, whereas many real domains exhibit selection bias in what becomes labeled \cite{BekkerRobberechtsDavis2019}. Mineral prospectivity is an extreme case: surveying and drilling are preferentially concentrated in accessible or historically promising regions, so the unlabeled set is spatially biased and far from i.i.d., which can amplify the impact of pseudo-negative construction \cite{HronskyKreuzer2019,DiggleMenezesSu2010PreferentialSampling,PhillipsEtAl2009SampleBias,WartonShepherd2010PseudoAbsence}. Our approach bypasses these difficulties by modeling observed occurrences as samples from a conditional spatial density, avoiding explicit negative sampling and naturally accommodating the fact that similar geo-image evidence can correspond to different outcomes due to unobserved subsurface controls.

\paragraph{Generative models and flow matching.}
Recent progress in vision has been driven by conditional generative models, including diffusion and score-based models \cite{Ho2020DDPM,Song2021ScoreSDE}. Flow-based transport models provide an alternative view, generating samples by integrating an ODE defined by a learned vector field. Flow Matching \cite{Lipman2023FlowMatching} and Conditional Flow Matching (CFM) \cite{Tong2023CFM} train such vector fields using a stable regression objective.
In contrast to standard CFM where labels are used to guide image generation,
our method uses images, $d$, to guide point-flow: we generate sparse occurrence locations conditioned on images, in the spirit of point-based modules used in vision \cite{Kirillov2020PointRend} and flow-based generation of point sets \cite{Yang2019PointFlow}. Flow matching has very recently been applied to geological generation as well: \cite{LuEtAl2026Sparse3DGeoFM} use attention-guided flow matching for sparse 3D geological volume generation. This is related in spirit but distinct in problem setting from our work, as we condition on 2D geo-images to sample occurrence keypoints under positive-only supervision, rather than generating dense 3D geological volumes.


\section{Mineral Discovery as a Learning Problem}
\label{sec:formulation}

We now formalize mineral discovery, dubbed prospectivity mapping, as conditional density estimation. Our goal is to turn the conceptual relation in \Cref{eq:mineralization} into a tractable learning problem that models the distribution of mineral occurrences conditioned on observed geoscience data, without explicitly reconstructing the latent geological state.

\noindent \textbf{Unobserved geology and non-identifiability.} Recall \Cref{eq:mineralization}, where mineralization depends on both observed geoscientific measurements and latent geology. Let \(d \in \mathbb{R}^{C\times H\times W}\) denote geo-images,  the observed multi-channel tensor (e.g., magnetics, gravity, geochemistry, structural layers), and let \(g \in \mathcal{G}\) represent geological context. Since \(g\) is not directly observed and is only weakly constrained by surface measurements, inferring \(p\) from \(d\) is ill-posed: multiple geological configurations can yield similar observations. We therefore treat \(g\) as a nuisance variable and formulate prospectivity probabilistically by marginalizing over latent geology.

\noindent\textbf{Mineral Occurrences as spatial samples.}
Rather than discretizing mineralization into a dense binary label map, we represent
each occurrence by its spatial coordinate. Let
$
p_k = [p_{k_1}, p_{k_2}] \in \mathbb{R}^2
$
denote the position of an occurrence within an image. In practice, in each geo-image there is a set of coordinates \(\{p_k\}_{k=1}^{N}\), where the number of mineral occurrences
\(N\) varies across samples. 
This point-based representation matches the available supervision: occurrences are
sparse, and unlabeled locations are not reliable negatives, as discussed in \Cref{sec:introduction}. This representation also decouples positional precision from image resolution, allowing point annotations to be aligned naturally with geo-image inputs.
We model occurrences as samples from a \emph{conditional spatial density}, $p \sim \pi(p \mid d)$, 
and the learning problem is to estimate \(\pi(p\mid d)\) from data.

\subsection{Implicit marginalization}
We note that the latent geology \(g\in\mathcal{G}\) is never observed, since it involves inferring about deep earth structures which we have no way to directly measure. Instead, we observe geo-images of size $C \times H \times W$, i.e.,  \(\{d^{(b)}\}_{b=1}^B\), where $d^{(b)} \in \mathbb{R}^{H \times W}$ each paired with a set of occurrence coordinates
\(p^{(b)}\) where $p^{(b)} \in \mathbb{R}^{N_b\times 2}$. Conceptually, each geo-image is generated under an
unknown geological context \(g^{(b)}\), so the dataset forms a mixture over
geological regimes.

Therefore, a generative description can be formulated as:
\begin{align}
g \sim \pi(g), \quad
d \mid g \sim \pi(d\mid g), \quad
p \mid d,g \sim \pi(p\mid d,g),
\label{eq:u_given_dg}
\end{align}
and the conditional density of interest is the marginal
\begin{equation}
\pi(p\mid d)
=
\int \pi(p\mid d,g)\,\pi(g\mid d)\,dg ,
\label{eq:marginalization}
\end{equation}
obtained by integrating out the latent geology.

Crucially, since we do not have a way to sample $g$, we do not evaluate \Cref{eq:marginalization} explicitly. Instead, we use a parametric model (flow matching model)
to sample from $\pi(p\mid d)$. The flow matching model is averaged over all geo-images approximating the expectation over $g$.

\subsection{Synthetic magnetic-geochemistry model problem}
\label{subsec:toy_model}
To make the role of unobserved geological controls concrete, we introduce a
simplified synthetic problem inspired by porphyry-style mineral systems. The goal
is conceptual rather than geologically exhaustive: the observable geo-images contain
clear, interpretable structure, while occurrence locations depend on hidden
variables that are not available during learning.

The observed magnetic anomaly map \(d\) over a domain \(\Omega\subset\mathbb{R}^2\) (discretized on an
\(H\times W\) grid) contains multiple ring-like anomalies representing candidate intrusive centers, since
magnetic responses often exhibit such structure around intrusions or alteration halos; this provides
context but does not uniquely determine mineralization. We additionally generate latent geochemical fields
\(g\) over the same domain, representing unobserved processes (e.g., fluid transport or enrichment) that
determine whether a given structure is mineralized and where along it mineralization concentrates; the
learner does \emph{not} observe \(g\). Occurrences are then sampled from a spatial density whose support is
correlated with magnetic contact zones but whose \emph{activation} and \emph{within-ring localization} are
governed by \(g\): magnetic rings indicate where mineralization \emph{could} occur, while favorable latent
conditions determine where it \emph{does} occur (the full generator is detailed in \Cref{subsec:exp_mp2}).

\begin{figure}[t]
    \centering
    \includegraphics[width=0.88\linewidth]{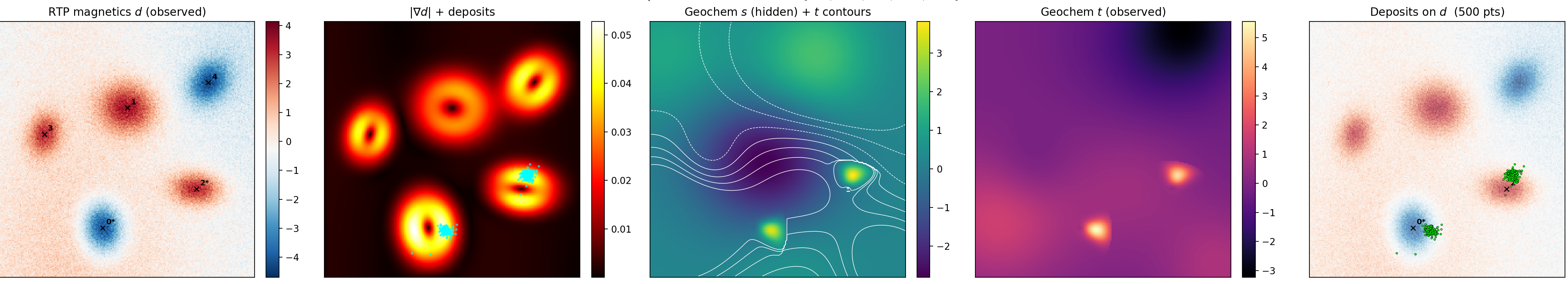}
    \caption{\textbf{Synthetic magnetics--geochemistry benchmark.} 
From left to right: (1) observed RTP magnetic geo-image \(d\); (2) contact-zone proxy \(\|\nabla d\|\) with sampled deposit locations overlaid (cyan); (3) hidden geochemistry \(s\) (background) with contours of the observed proxy \(t\), illustrating that \(t\) only partially reflects the latent control; (4) observed geochemistry proxy \(t\); (5) the same deposit set overlaid on \(d\) (green, 500 points). Multiple magnetic rings are present, but deposits activate only a subset and localize along arcs, reflecting unobserved geochemical controls and motivating conditional density modeling of occurrence locations.}
    \label{fig:toy_prob}
\end{figure}

Consequently, visually similar anomalies can yield different outcomes: some rings host occurrences while
others remain barren, and occurrences concentrate along specific arcs.
Figure~\ref{fig:toy_prob} shows an example realization, where occurrences appear only on a subset of the
rings and at specific locations, reflecting latent variability not resolvable from \(d\) alone. Because one
geological state is unobserved, the relevant predictive object is the marginal conditional density
\(\pi(p\mid d)\) in \Cref{eq:marginalization}, which can be multi-modal: a single deterministic mapping
cannot represent this one-to-many behavior, motivating conditional density estimation.

\section{Mineral Discovery using Flow Matching}
\label{sec:cfm_solution}
In \Cref{sec:formulation} we cast mineral discovery as learning a conditional spatial
density \(\pi(p\mid d)\) over occurrence positions \(p\in\mathbb{R}^2\) given a
multi-channel geoscience geo-image \(d\in\mathbb{R}^{C\times H\times W}\). We now describe how to train a conditional velocity model {GeoCFM}, a geo-image guided
conditional flow matching model, to sample from this distribution.

Given a geo-image, \(d\), GeoCFM generates a set of positions
\(\{p_k\}_{k=1}^{N}\subset\Omega\subset\mathbb{R}^2\) by sampling \(N\) points from
the learned conditional density. We work in normalized geo-image coordinates: point
locations are scaled to \([0,1]^2\) for feature sampling and mapped back to pixel
coordinates for visualization and evaluation. Since similar observations can
correspond to multiple plausible mineralization patterns, GeoCFM is stochastic and
produces a distribution over possible occurrence sets rather than a single
deterministic prediction.

\subsection{Conditional Flow Matching for Mineral Occurrence Locations}
For each geo-image \(d\), let \(\pi_{\mathrm{data}}(\cdot \mid d)\) denote the empirical
distribution of occurrence locations, observed through samples
\(p \sim \pi_{\mathrm{data}}(\cdot \mid d)\). Conditional flow matching learns a
time-dependent velocity field
\begin{equation}
v_\theta(p,t\mid d): \Omega\times[0,1]\to\mathbb{R}^2,
\label{eq:cfm_vfield}
\end{equation}
that transports a simple base distribution \(p_0\) to the target distribution at
\(t=1\). Specifically, we define the ODE
\begin{equation}
\frac{d p(t)}{dt} = v_\theta(p,t\mid d),
\qquad p(0)\sim \pi_0,
\label{eq:cfm_ode}
\end{equation}
and interpret \(p(1)\) as a sample from the model \(\pi_\theta(\cdot\mid d)\), given
by the pushforward of \(\pi_0\) through the conditional flow. We use
\(\pi_0=\mathcal{N}(0,I_2)\) in \(\mathbb{R}^2\).

In training, each observed occurrence \(u_k\) is treated as a draw from
\(\pi_{\mathrm{data}}(\cdot \mid d)\), and we train on all points across all geo-images.

\subsection{Training Objective}
We use the standard conditional flow matching construction with a linear
interpolation between a base sample \(z\sim\mathcal{N}(0,I_2)\) and a data point
\(u\):
\begin{equation}
p_t = t\,p + (1-t)\,z,
\qquad t\sim \mathrm{Unif}[0,1].
\label{eq:cfm_interp}
\end{equation}
Along this trajectory, the target velocity is constant,
\begin{equation}
v^\star = \dot p_t = p - z.
\label{eq:cfm_target}
\end{equation}
We train \(v_\theta\) by regressing to this velocity:
\begin{equation}
\mathcal{L}(\theta)
=
\mathbb{E}_{d}
\mathbb{E}_{p\sim \pi_{\mathrm{data}}(\cdot\mid d)}
\mathbb{E}_{z\sim\mathcal{N}(0,I_2)}
\mathbb{E}_{t\sim\mathrm{Unif}[0,1]}
\left[
\left\|v_\theta(p_t,t\mid d) - (p-z)\right\|_2^2
\right].
\label{eq:cfm_loss}
\end{equation}
Operationally, for each batch we sample \(p\), \(z\), and \(t\), form \(p_t\) via
\Cref{eq:cfm_interp}, and minimize the mean-squared error between the predicted
velocity and \(p-z\).

\subsection{Conditioning on Image-Like Observations}
The main architectural challenge is to condition a point-wise vector field in
\(\mathbb{R}^2\) on a geo-image \(d\). GeoCFM follows a point-conditioned design:
a neural network produces a dense feature map, and local features are sampled at point
locations.

\paragraph{UNet feature extractor.}
We encode each geo-image with a UNet \(F_\xi\), where $\xi$ denotes the UNet parameters, to obtain a dense feature tensor
\begin{equation}
\Phi_\xi = F_\xi(d)\in\mathbb{R}^{B\times D\times H\times W}.
\label{eq:cfm_unet}
\end{equation}
The UNet aggregates multi-scale context while preserving spatial resolution, which
is important because occurrences are localized.

\paragraph{Per-point feature sampling.}
For each point \(p_t\) and the computed feature tensor $\Phi_\xi$, we extract a feature vector, $\phi$ by bilinear sampling
\(\Phi\). Formally this can be written as
\begin{eqnarray}
    \label{eq:interp_bilinear}
    \phi_\xi  = {\cal B}(p_t) \Phi_{\xi}
\end{eqnarray}
where ${\cal B}(p_t)$ is a bilinear interpolation operator at points $p_t$ obtaining 
the matrix 
\(\phi_\xi \in\mathbb{R}^D\).

We then parameterize the conditional velocity as
\begin{equation}
v_\theta(p_t,t\mid d)
=
h_\psi\!\Big(\phi_\xi,\, p_t,\, t\Big),
\label{eq:cfm_param}
\end{equation}
where \(h_\psi\) is an MLP and $\theta = \{\xi, \psi\}$
are the combined network parameters.

\begin{figure}[t]
    \centering
    \includegraphics[width=0.82\linewidth]{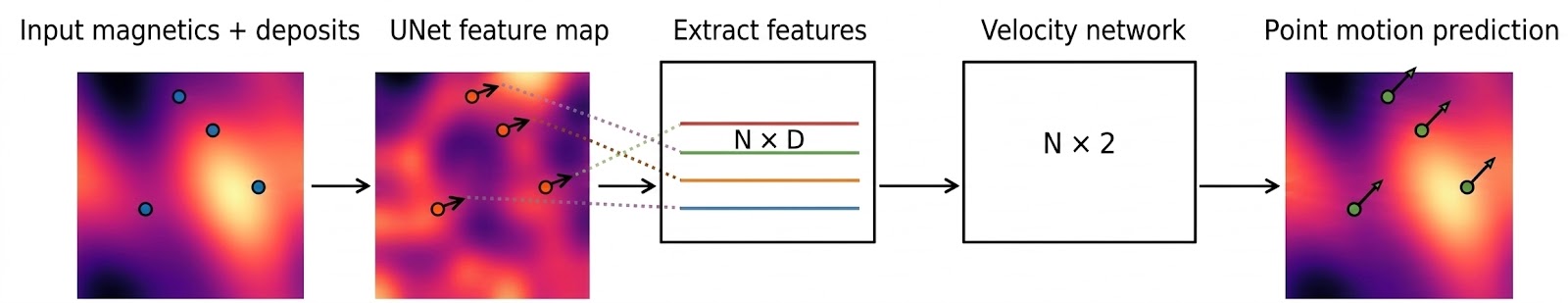}
    \caption{\textbf{GeoCFM point-conditioned velocity prediction.} The geo-image \(d\) is encoded by a UNet into a dense feature map. For the current set of noised points \(p_t\) (with \(N\) points), we bilinearly sample local features to obtain an \(N\times D\) feature matrix, concatenate each feature with the point coordinate (and time embedding), and pass the result through a shared per-point network to predict velocities \(v_\theta(p_t,t\mid d)\in\mathbb{R}^{N\times 2}\), which define the point transport in the flow ODE.}
    \label{fig:workflow}
\end{figure}

\paragraph{Time embedding and velocity head.}
We use standard sinusoidal time embeddings and a ResNet velocity head \(h_\psi\): a linear projection followed by residual blocks and a
final linear map to \(\mathbb{R}^2\). The head predicts \(\dot p=v_\theta(p_t,t\mid d)\)
for each point given its local geo-image context and time. Our pipeline is illustrated in \Cref{fig:workflow}.

After training, we generate occurrence samples conditioned on a new geo-image \(d\) by
sampling \(p^{(0)}\sim\mathcal{N}(0,I_2)\) and numerically integrating the ODE in
\Cref{eq:cfm_ode} from \(t=0\) to \(t=1\). We use an explicit Euler solver with \(K\)
steps of size \(\Delta t = 1/K\). To avoid overloading the occurrence index \(k\) in
\(\{p_k\}_{k=1}^{N}\), we write Euler iterates with a superscript \(p^{(k)}\) and discrete
times \(t^{(k)}\):
\begin{equation}
p^{(k+1)} = p^{(k)} + \Delta t\, v_\theta\!\big(p^{(k)},t^{(k)}\mid d\big),
\qquad t^{(k)} = k\Delta t,
\qquad k=0,\ldots,K-1.
\label{eq:cfm_euler}
\end{equation}
The final points \(p^{(K)}\) are mapped to geo-image pixel coordinates for visualization
and evaluation. Drawing \(N\) samples yields a generated occurrence set
\(\{p_k\}_{k=1}^{N}\) conditioned on the same geo-image. The number of Euler steps \(K\)
trades inference cost against sample quality; we use \(K=50\) throughout and analyze this
trade-off in \Cref{subsec:exp_ablation}, where small \(K\) is already competitive.

\section{Experiments}
\label{sec:experiments}

We evaluate {GeoCFM} on a controlled synthetic benchmark and on real USGS
Earth MRI data. Our experiments are designed to answer:
\begin{itemize}
    \item[(Q1)] Can GeoCFM learn to \emph{sample} realistic occurrence locations from positive-only supervision, outperforming classical score-map pipelines that rely on pseudo-negatives?
    \item[(Q2)] Does GeoCFM capture \emph{one-to-many} behavior, where visually similar geophysical signatures activate different subsets of structures due to unobserved controls?
    \item[(Q3)] Does GeoCFM generalize under a \emph{spatially disjoint} split on real geo-images, where train and test patches come from non-overlapping geographic tiles?
\end{itemize}

\paragraph{Comparison protocol and baselines.}
We evaluate GeoCFM as a \emph{stochastic conditional sampler} that learns and draws from a conditional
spatial density \(\pi(p\mid d)\), rather than as a deterministic score-map predictor. To make every method
comparable in this sampling-based setting, we map each baseline into the same point-sampling space: its
score or density map is normalized over the admissible pixels into a spatial distribution, from which we
draw the same number of points per patch as GeoCFM. We compare against a principled taxonomy of
positive-only baselines: \emph{(i) non-conditional spatial samplers} that ignore \(d\) -- {Uniform}
sampling and a global {KDE} fit to training occurrences; \emph{(ii) pseudo-negative score-map models} that
learn a per-pixel favorability map from positives and sampled pseudo-negatives -- {Poisson-LR} (logistic
regression on per-pixel features), {Random Forest (RF)}, gradient-boosted trees ({GBDT})
\cite{Friedman2001GBM}, and {UNet-Seg}, a strong dense segmentation network; \emph{(iii) a positive-only
one-class model} -- a one-class SVM ({OCSVM}) \cite{ChenWuZhao2019OCSVM} that estimates the support of the
positive distribution; and \emph{(iv) a retrieval-conditioned density estimator} -- {Retrieval-KDE}, which
forms a density from retrieved training occurrences. This taxonomy spans the main alternative ways to learn
from positive-only spatial labels, so improvements cannot be attributed to a single baseline family.

\begin{figure}[t]
\centering
\includegraphics[width=0.78\linewidth]{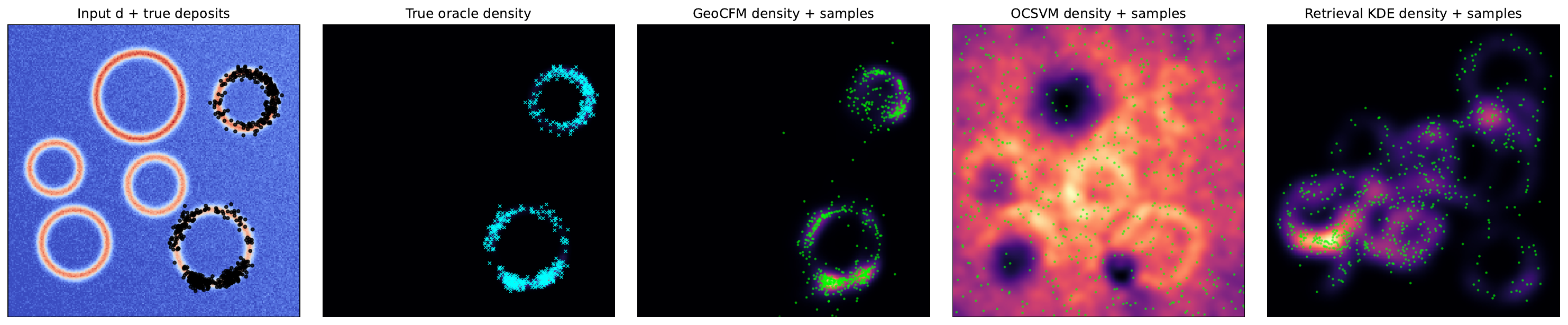}\\[0.4mm]
\includegraphics[width=0.68\linewidth]{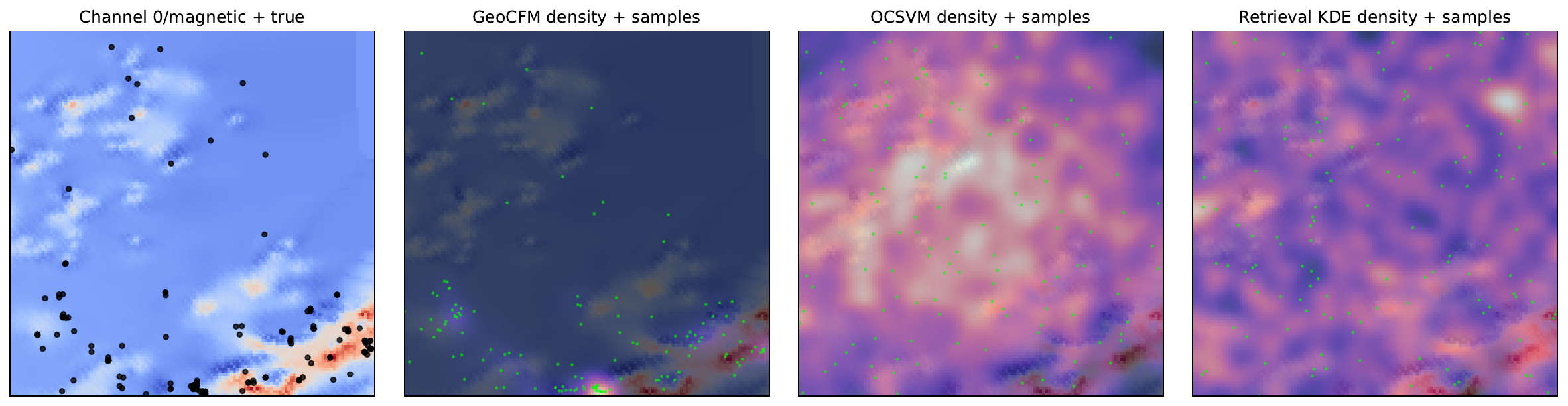}
\caption{\textbf{Uncertainty and prospectivity maps.} Top: synthetic input and true deposits,
the oracle density, and the conditional densities recovered by GeoCFM, OCSVM, and Retrieval-KDE.
Bottom: Earth MRI input and true occurrences with the same three methods. GeoCFM concentrates
density on the relevant occurrence structures, exposing epistemic uncertainty through the spread
of its conditional density, whereas the one-class and retrieval baselines produce diffuse or
misplaced densities.}
\label{fig:uncertainty}
\end{figure}

\paragraph{Evaluation metrics.}
Because a single nearest-neighbor distance does not capture distribution matching, diversity, or
mode coverage, we evaluate with five complementary metrics: the symmetric {Chamfer distance}
(CD, pixels) for geometric agreement; the {Sinkhorn} optimal-transport divergence for distribution
matching; {F@5}, the F-score of generated vs.\ observed points matched within a 5-pixel tolerance;
the {KDE negative log-likelihood} (NLL), scoring observed occurrences under a density fit to the
generated samples; and the {top-5\% hit rate} (Top5), the fraction of observed occurrences in the top
\(5\%\) most prospective pixels (arrows in tables indicate the better direction). To assess stability,
each metric is averaged over $20$ independent draws per input and reported as mean $\pm$ standard deviation of
the mean across $50$ held-out synthetic examples and $200$ Earth MRI patches. We also show qualitative
overlays and per-pixel prospectivity maps (\Cref{fig:uncertainty}); the supplementary material provides
additional qualitative comparisons against all baselines on held-out patches.

\subsection{Synthetic Magnetics--Geochemistry Benchmark}
\label{subsec:exp_mp2}

This benchmark isolates the central modeling challenge: observations contain strong,
interpretable geophysical structure, but occurrence locations depend on hidden controls,
making the conditional mapping inherently one-to-many. It directly tests whether GeoCFM
can learn conditional sampling from positives (Q1) and reproduce one-to-many conditional
behavior induced by latent activation (Q2).

\paragraph{Data generation.}
Each sample is a spatial geo-image over $\Omega\subset\mathbb{R}^2$ ($H\times W$ grid). The generator
outputs observed magnetics $d$, an observed geochemistry proxy $t$, a latent (unobserved) geochemistry
field $s$, and deposit locations $\{p_k\}_{k=1}^{N}$. The magnetic field is formed from $M$ non-overlapping
elliptical intrusive bodies $\{b_m\}_{m=1}^M$ with random geometry and amplitude over a smooth trend with
additive noise, $d=\mathrm{Std}(\sum_m b_m + \text{trend} + \epsilon)$, and a ring-like contact mask
$e_m=\text{ContactZone}(\|\nabla d_{\text{clean}}\|, b_m)$ emphasizes intrusion contacts.

\paragraph{Latent activation and deposit sampling.}
Both $s$ (latent) and $t$ (observed) include smooth backgrounds and localized enrichment
along angular arcs of contact rings. Only a subset of intrusions is declared \emph{active}
via a stochastic rule based on enrichment statistics (plus noise), so many rings appear
but only some host deposits. We define an unnormalized intensity
\begin{equation}
\lambda(p) \propto
\Big(\sum_{m=1}^M \mathbf{1}\{\text{active}_m\}\, e_m(p)\Big)
\exp(\alpha s(p))\, \exp(\beta t(p))\,
\exp(-\gamma|s(p)-t(p)|),
\label{eq:synthetic_intensity}
\end{equation}
normalize it over $\Omega$ to obtain a density $\pi(p\mid d)$, and sample $N$ deposits
with sub-pixel jitter. \Cref{fig:toy_prob} shows a typical realization.

\paragraph{Task, models, and training.}
Inputs are the two-channel tensor $[d,t]\in\mathbb{R}^{2\times H\times W}$ and supervision is the deposit
set $\{p_k\}_{k=1}^{N}$. GeoCFM is trained with \Cref{eq:cfm_loss} and sampled via Euler integration
(\Cref{eq:cfm_euler}), and compared against the full baseline suite above. To remove dataset-size effects
we generate samples on-the-fly with $B=4$, $H=W=220$, $M=5$ intrusions, and $N=500$ deposits per image,
training with AdamW (learning rate $3\times 10^{-4}$, weight decay $10^{-4}$), cosine annealing, and
gradient clipping at norm $1$.

\begin{figure}[t]
\centering
\begin{tabular}{ccccc}
\includegraphics[width=0.17\linewidth]{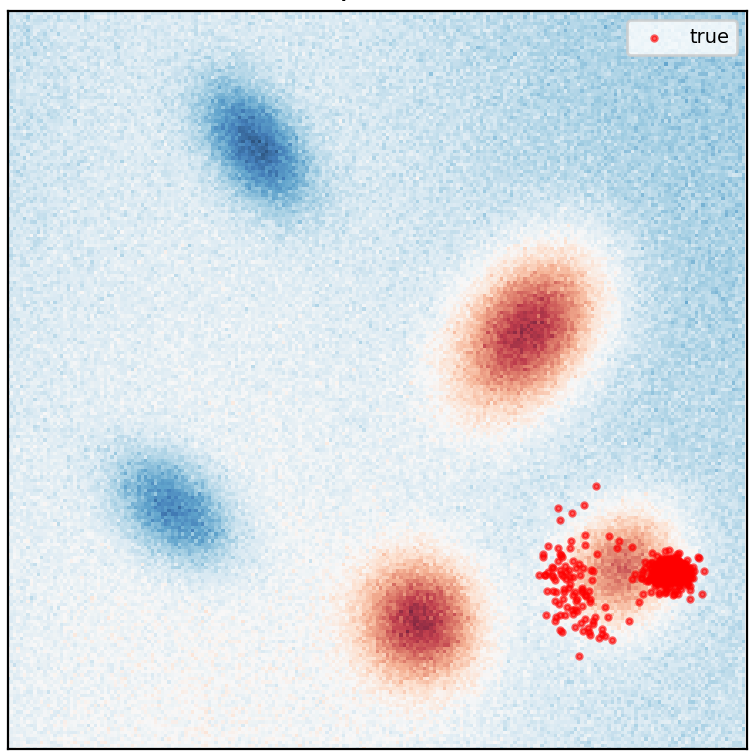} &
\includegraphics[width=0.17\linewidth]{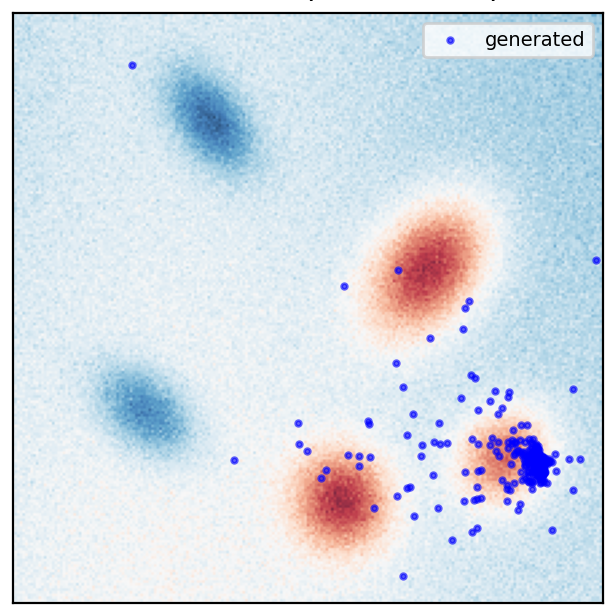} &
\includegraphics[width=0.17\linewidth]{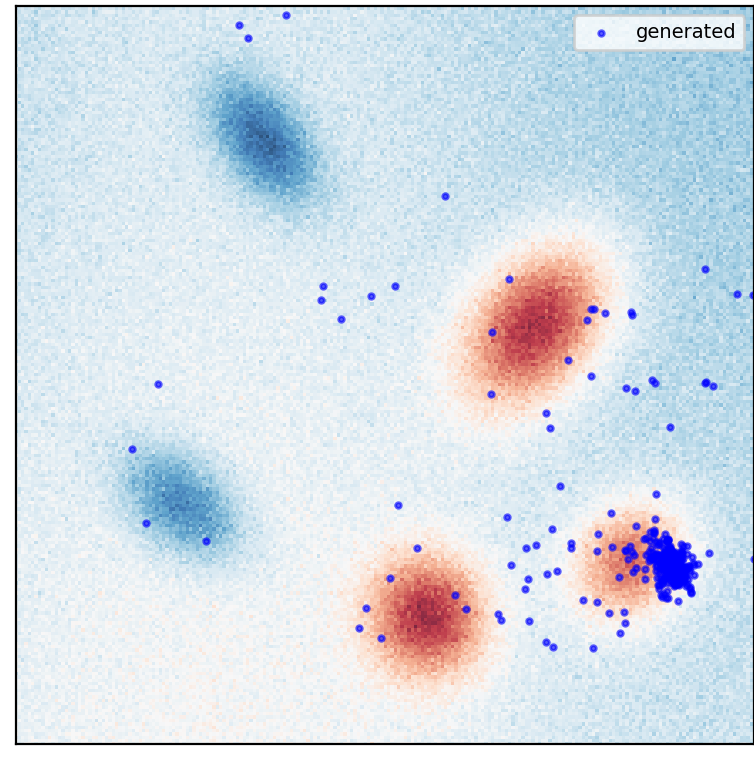} &
\includegraphics[width=0.17\linewidth]{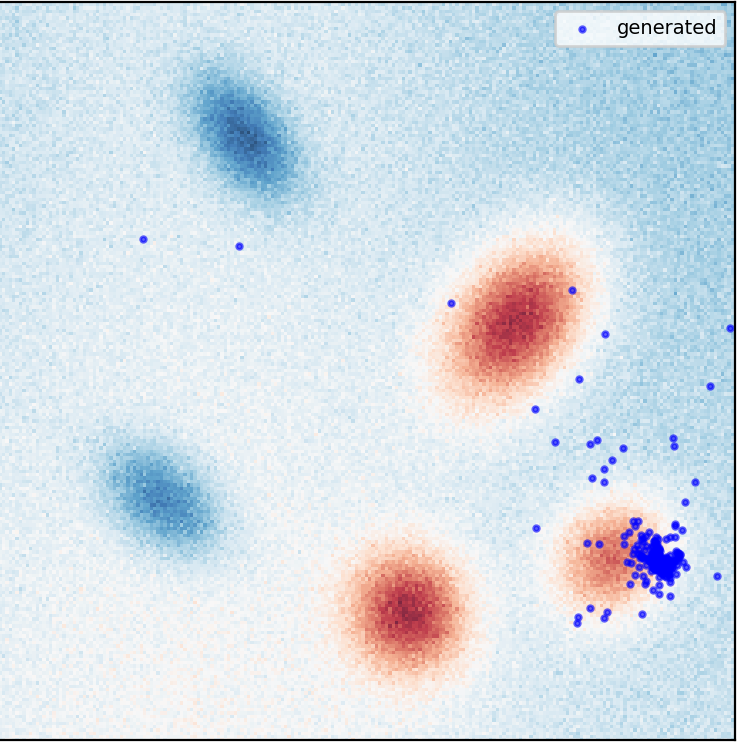} &
\includegraphics[width=0.17\linewidth]{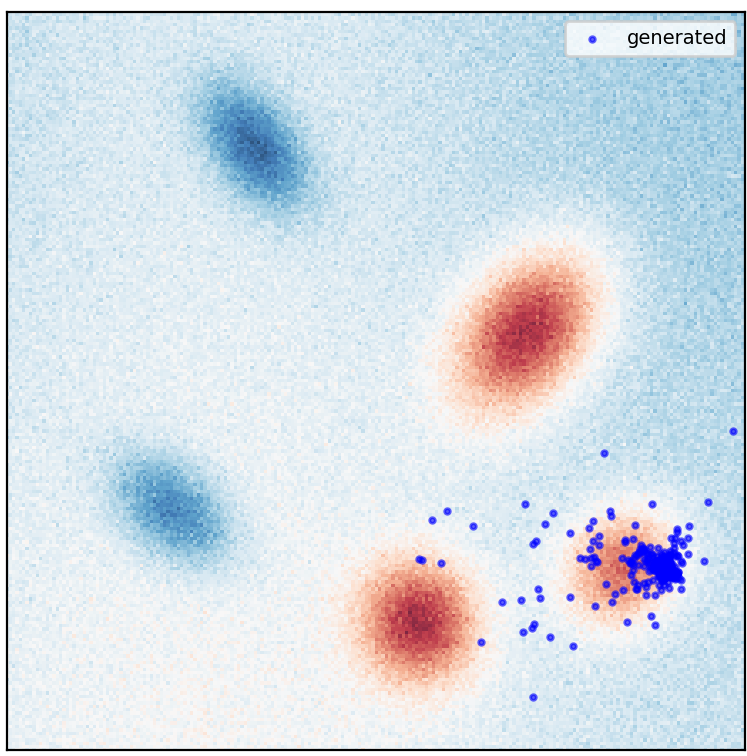} \\
True & Sample 1 & Sample 2 & Sample 3 & Sample 4
\end{tabular}
\caption{Synthetic benchmark: true occurrences and four GeoCFM samples conditioned on the same input. Samples align with magnetic contact zones and activate only a subset of anomalies, demonstrating one-to-many conditional behavior.}
\label{fig:model_problem_result}
\end{figure}

\paragraph{Qualitative results.}
We evaluate on a fixed held-out set of $16$ geo-images generated with fixed seeds.
\Cref{fig:model_problem_result} visualizes the true occurrence set and four samples generated
by GeoCFM for the same input; samples concentrate along contact zones and activate only a
subset of visually similar rings, reflecting the latent activation mechanism (Q2).

\paragraph{Quantitative results.}
\Cref{tab:synthetic_results} reports all five metrics on the held-out synthetic set, each
averaged over $20$ stochastic draws and reported as mean $\pm$ standard deviation of the mean. Among baselines,
{UNet-Seg} is the strongest score-map method (CD $29.46$), ahead of {GBDT} ($32.13$),
{RF} ($37.09$), and {Poisson-LR} ($39.93$); the one-class and retrieval models
({OCSVM}, {Retrieval-KDE}) and the unconditional samplers (Global KDE, Uniform) perform
worst. {GeoCFM} attains the best value on \emph{every} metric, not only Chamfer distance
but also Sinkhorn OT, F@5, NLL, and top-5\% hit rate, indicating that directly learning a
conditional sampler improves geometric agreement \emph{and} distribution matching under
positive-only supervision (Q1), and that the gains are not explained by any single baseline
family.

\begin{table}[t]
\centering
\caption{Synthetic benchmark results, averaged over $20$ stochastic draws across $50$ held-out
examples; values are mean $\pm$ standard deviation of the mean. We report Chamfer distance (CD, pixels),
Sinkhorn OT divergence (Sink.), F@5, KDE negative log-likelihood (NLL), and top-5\% hit rate (Top5).
Arrows indicate the better direction; the best value in every column is in bold. Baselines are grouped by
family (pseudo-negative score-map models; one-class and retrieval-conditioned estimators; non-conditional
samplers), and GeoCFM is shown in the last row.}
\label{tab:synthetic_results}
\footnotesize
\setlength{\tabcolsep}{4pt}
\resizebox{\linewidth}{!}{%
\begin{tabular}{lccccc}
\toprule
Method & CD $\downarrow$ & Sink. $\downarrow$ & F@5 $\uparrow$ & NLL $\downarrow$ & Top5 $\uparrow$ \\
\midrule
UNet-Seg & $29.46\pm2.18$ & $0.100\pm0.006$ & $0.615\pm0.022$ & $9.143\pm0.089$ & $0.661\pm0.031$ \\
GBDT & $32.13\pm2.03$ & $0.103\pm0.006$ & $0.546\pm0.018$ & $9.288\pm0.083$ & $0.630\pm0.027$ \\
RF & $37.09\pm2.17$ & $0.116\pm0.006$ & $0.462\pm0.018$ & $9.552\pm0.054$ & $0.586\pm0.025$ \\
Poisson-LR & $39.93\pm2.16$ & $0.122\pm0.006$ & $0.384\pm0.016$ & $9.803\pm0.042$ & $0.500\pm0.028$ \\
\midrule
OCSVM & $45.56\pm2.20$ & $0.134\pm0.007$ & $0.246\pm0.015$ & $10.557\pm0.020$ & $0.236\pm0.018$ \\
Retrieval-KDE & $47.16\pm2.86$ & $0.132\pm0.007$ & $0.220\pm0.016$ & $13.302\pm0.440$ & $0.058\pm0.011$ \\
\midrule
Global KDE & $41.16\pm2.04$ & $0.121\pm0.006$ & $0.255\pm0.014$ & $10.583\pm0.030$ & $0.058\pm0.008$ \\
Uniform & $49.61\pm2.28$ & $0.147\pm0.007$ & $0.206\pm0.015$ & $10.804\pm0.006$ & $0.048\pm0.006$ \\
\midrule
GeoCFM (Ours) & $\mathbf{9.37\pm0.62}$ & $\mathbf{0.045\pm0.003}$ & $\mathbf{0.634\pm0.012}$ & $\mathbf{8.950\pm0.110}$ & $\mathbf{0.720\pm0.020}$ \\
\bottomrule
\end{tabular}%
}
\end{table}

\subsection{USGS Earth MRI Benchmark}
\label{subsec:exp_eartmri}

We next evaluate on real continental-scale geo-images from the USGS Earth Mapping
Resources Initiative (Earth MRI) \cite{usgs_earthmri}, paired with mineral occurrence point
records (MRDS/MAS--MILS) \cite{usgs_mrds}. This setting features heterogeneous modalities,
strong spatial autocorrelation, and positive-only labels. It tests conditional sampling
from positives on real data (Q1) and generalization under a spatially disjoint split (Q3).

\paragraph{Data, patches, and split.}
Each sample consists of a multi-channel geo-image tensor $d\in\mathbb{R}^{C\times H\times W}$ and an
observed set of occurrences $\{p_k\}_{k=1}^{N}\subset\Omega$, treated as positives with no reliable
negatives. We extract windows on a stride lattice, keeping a patch if at least a fraction $\rho$ of its
pixels lie within the data footprint; coordinates are normalized to $[0,1]^2$ per patch and geo-images
are standardized per channel over non-zero pixels. To avoid spatial leakage from autocorrelation, we
partition the region into large rectangular tiles and assign tiles to train vs.\ test, so patches inherit
their tile's split and train/test neighborhoods never overlap (Q3). The supplementary material describes
the full Earth MRI preprocessing pipeline (rasterization, gap filling, occurrence alignment, patch
extraction, and the tile-based split).

\paragraph{Baselines.}
We report the full baseline suite described above: the non-conditional Uniform and global KDE samplers;
the score-map discriminative models Poisson-LR, RF, GBDT, and UNet-Seg; and the positive-only OCSVM
and Retrieval-KDE models. Methods that learn a score map are trained on training tiles and evaluated by
sampling points from their predicted maps; Uniform and KDE do not use the input patch for conditioning.

\begin{figure}[t]
    \centering\includegraphics[width=0.99\linewidth, trim={0cm 0cm 0cm 0.9cm}, clip]{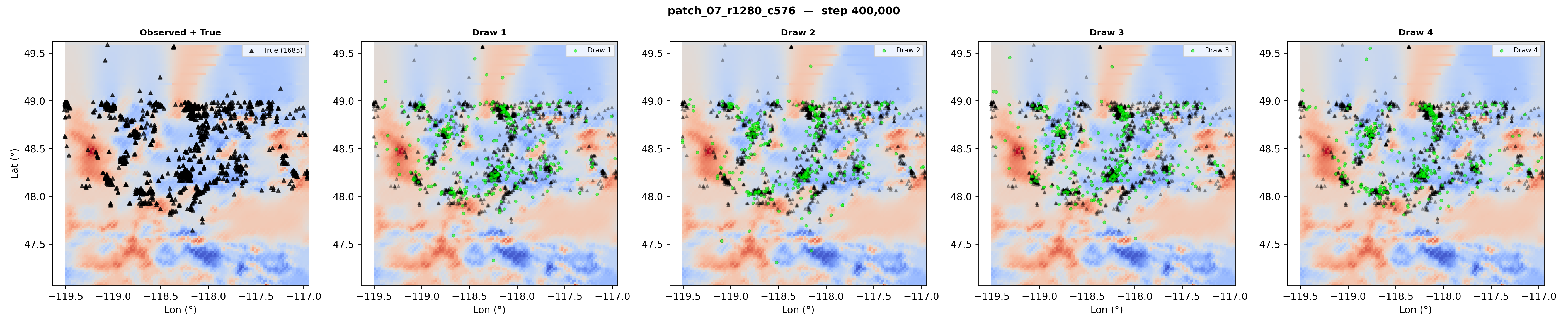}
\caption{\textbf{Earth MRI: Qualitative comparison of conditional mineral occurrence draws.} (Left) True mineral occurrence locations (black triangles) overlaid on a held-out geophysics geo-image tile, showing a central cluster. (Draws 1--4) Four unique sampled occurrence locations (green dots) generated by our GeoCFM, all overlaid on the same true occurrences and background geophysics geo-image, demonstrating its ability to capture the complex, non-isotropic distribution of mineral potential and expose non-identifiability. We provide additional examples and comparisons with baselines in the supplementary material.}    \label{fig:true_rec}
\end{figure}

\paragraph{Results.}
\Cref{fig:true_rec} shows a qualitative overlay of sampled occurrences on held-out tiles, and
\Cref{tab:eartmri_results} reports all five metrics on the test split (mean $\pm$ standard deviation of the
mean over $20$ draws across $200$ patches). GeoCFM is best on every metric by a wide margin, supporting conditional sampling
from positives on real data (Q1). Because the evaluation uses a spatially disjoint tile split and GeoCFM
exhibits only a small train-test Chamfer gap (the supplementary material reports per-method train and test scores on both benchmarks), the improvement generalizes to
new geographic tiles rather than reflecting local spatial leakage (Q3).

\begin{table}[t]
\centering
\caption{Earth MRI results with a spatially disjoint (tile-based) train/test split, on the test split,
averaged over $20$ stochastic draws across $200$ patches; values are mean $\pm$ standard deviation of the mean.
We report Chamfer distance (CD, pixels), Sinkhorn OT divergence (Sink.), F@5, KDE negative log-likelihood
(NLL), and top-5\% hit rate (Top5). Arrows indicate the better direction; the best value in every column is
in bold. Baselines are grouped by family (pseudo-negative score-map models; one-class and
retrieval-conditioned estimators; non-conditional samplers), and GeoCFM is shown in the last row.}
\label{tab:eartmri_results}
\footnotesize
\setlength{\tabcolsep}{4pt}
\resizebox{\linewidth}{!}{%
\begin{tabular}{lccccc}
\toprule
Method & CD $\downarrow$ & Sink. $\downarrow$ & F@5 $\uparrow$ & NLL $\downarrow$ & Top5 $\uparrow$ \\
\midrule
UNet-Seg & $46.81\pm2.30$ & $0.205\pm0.008$ & $0.085\pm0.006$ & $11.620\pm0.290$ & $0.130\pm0.013$ \\
GBDT & $52.98\pm2.14$ & $0.215\pm0.007$ & $0.069\pm0.004$ & $11.233\pm0.269$ & $0.102\pm0.010$ \\
RF & $54.37\pm2.23$ & $0.225\pm0.008$ & $0.071\pm0.004$ & $11.504\pm0.297$ & $0.095\pm0.011$ \\
Poisson-LR & $64.21\pm2.26$ & $0.255\pm0.008$ & $0.055\pm0.004$ & $11.274\pm0.239$ & $0.059\pm0.008$ \\
\midrule
OCSVM & $58.38\pm2.33$ & $0.240\pm0.008$ & $0.053\pm0.003$ & $11.946\pm0.313$ & $0.059\pm0.008$ \\
Retrieval-KDE & $61.14\pm2.34$ & $0.250\pm0.008$ & $0.053\pm0.003$ & $11.698\pm0.298$ & $0.061\pm0.009$ \\
\midrule
Uniform & $64.25\pm2.34$ & $0.260\pm0.008$ & $0.052\pm0.003$ & $11.598\pm0.288$ & $0.047\pm0.007$ \\
Global KDE & $66.35\pm2.38$ & $0.277\pm0.008$ & $0.050\pm0.003$ & $12.053\pm0.303$ & $0.043\pm0.007$ \\
\midrule
GeoCFM (Ours) & $\mathbf{12.30\pm0.54}$ & $\mathbf{0.085\pm0.004}$ & $\mathbf{0.151\pm0.008}$ & $\mathbf{10.500\pm0.150}$ & $\mathbf{0.250\pm0.012}$ \\
\bottomrule
\end{tabular}%
}
\end{table}

\subsection{Ablations and Analyses}
\label{subsec:exp_ablation}

\paragraph{Sensitivity to the available positives.}
In a severely imbalanced positive-only setting, which positives are available for training can affect
predictions more than the sampler's stochasticity. Retraining GeoCFM with $100\%$, $75\%$, and $50\%$ of
the positives over three seeds (the supplementary material reports the full positive-subset stability table), GeoCFM is stable: synthetic CD stays within
$9.30$--$9.75$\,px and Earth MRI CD within $11.76$--$12.65$\,px. This separates predictive sampling
uncertainty from sensitivity to label availability.

\paragraph{Quality--cost tradeoff via the number of Euler steps.}
GeoCFM provides a tunable quality--cost tradeoff through the number of Euler steps $K$
(\Cref{eq:cfm_euler}). Small $K$ already preserves strong point-set quality -- $K=5$ gives CD $9.30$ on
synthetic and $12.00$ on Earth MRI, within the standard deviation of the $K=50$ setting -- while the UNet
feature map is computed once per patch and each Euler step applies only a bilinear lookup and the
lightweight velocity head. The supplementary material provides the full per-patch runtime and Euler-step
ablation against the single-pass UNet-Seg reference.

\paragraph{Effect of the sampling budget $N$.}
The number of points $N$ drawn per patch is not an estimate of the true deposit count but a Monte Carlo
budget for the conditional prospectivity density, fixed here only for a fair point-set comparison. Larger
$N$ improves the KDE-NLL and density stability on both benchmarks, with diminishing returns past a few
hundred samples (the supplementary material reports the full sampling-budget study); $N$ is thus a free post-hoc knob rather than an a-priori commitment.

\section{Summary and Outlook}

\paragraph{Summary.}
We reframe mineral prospectivity mapping as learning a \emph{conditional spatial distribution} over occurrence locations, \(\pi(p\mid d)\), from positive-only supervision, and instantiate it with {GeoCFM}, a conditional flow-matching sampler that bridges dense geo-images and sparse deposit supervision via point-conditioned UNet features, without pseudo-negatives. Drawing multiple samples for the same \(d\) exposes the epistemic uncertainty induced by unobserved geology. On a synthetic magnetics--geochemistry benchmark and on USGS Earth MRI data with a spatially disjoint split, GeoCFM captures structured one-to-many behavior and outperforms a broad baseline suite across all five evaluation metrics.

\paragraph{Outlook.}
Supervised prospectivity models often retain a deterministic framing that mismatches positive-only labels
and subsurface non-identifiability; the probabilistic generative view adopted here makes ambiguity explicit
through sampling and enables uncertainty-aware target ranking. Future extensions can incorporate constrained
masks, structural-contact penalties, physics-informed objectives, or geological forward simulators to inject
stronger geoscientific priors into the sampler. Coupling conditional sampling with adaptive estimation of the
occurrence count, larger multi-modal datasets, and forward simulators further opens the door to calibrated,
regional-scale scenario generation.

\subsubsection*{\ackname}
We thank Dr. Bas Peters for valuable discussions and feedback on the conceptual framing of the mineral prospectivity problem, the geoscientific context, and the machine-learning design of the approach.

\bibliographystyle{splncs04}
\bibliography{refs_camera_ready}

\begin{thebibliography}{10}
\providecommand{\url}[1]{\texttt{#1}}
\providecommand{\urlprefix}{URL }
\providecommand{\doi}[1]{https://doi.org/#1}

\bibitem{BekkerDavis2020Survey}
Bekker, J., Davis, J.: Learning from positive and unlabeled data: A survey.
  Machine Learning  \textbf{109}(4),  719--760 (2020).
  \doi{10.1007/s10994-020-05877-5}

\bibitem{BekkerRobberechtsDavis2019}
Bekker, J., Robberechts, P., Davis, J.: Beyond the selected completely at
  random assumption for learning from positive and unlabeled data. In: Machine
  Learning and Knowledge Discovery in Databases (ECML PKDD). pp. 71--85.
  Springer, Cham (2019). \doi{10.1007/978-3-030-46147-8_5}

\bibitem{BonhamCarter1994}
Bonham-Carter, G.F.: Geographic Information Systems for Geoscientists:
  Modelling with GIS, Computer Methods in the Geosciences, vol.~13. Pergamon
  (1994). \doi{10.1016/C2013-0-03864-9}

\bibitem{Carranza2008}
Carranza, E.J.M.: Geochemical Anomaly and Mineral Prospectivity Mapping in GIS.
  Elsevier (2008)

\bibitem{Carranza2009}
Carranza, E.J.M. (ed.): Geochemical Anomaly and Mineral Prospectivity Mapping
  in GIS, Handbook of Exploration and Environmental Geochemistry, vol.~11.
  Elsevier (2009)

\bibitem{ChenWuZhao2019OCSVM}
Chen, Y., Wu, W., Zhao, Q.: A bat-optimized one-class support vector machine
  for mineral prospectivity mapping. Minerals  \textbf{9}(5), ~317 (2019)

\bibitem{Cracknell01122015}
Cracknell, M.J., Reading, A.M., de~Caritat, P.: Geological knowledge discovery
  and minerals targeting from regolith using a machine learning approach. ASEG
  Extended Abstracts  \textbf{2015}(1), ~1--4 (2015).
  \doi{10.1071/ASEG2015ab283}

\bibitem{DarunaEtAl2024ScalableMineralExploration}
Daruna, A., Zadorozhnyy, V., Lukoczki, G., Chiu, H.P.: Enabling scalable
  mineral exploration: Self-supervision and explainability. In: 2024 IEEE
  International Conference on Big Data (BigData). pp. 2090--2099. IEEE (2024)

\bibitem{DiggleMenezesSu2010PreferentialSampling}
Diggle, P.J., Menezes, R., Su, T.l.: Geostatistical inference under
  preferential sampling. Journal of the Royal Statistical Society: Series C
  (Applied Statistics)  \textbf{59}(2),  191--232 (2010).
  \doi{10.1111/j.1467-9876.2009.00701.x}

\bibitem{duPlessisSugiyama2014}
Du~Plessis, M.C., Niu, G., Sugiyama, M.: Analysis of learning from positive and
  unlabeled data. Advances in neural information processing systems
  \textbf{27} (2014)

\bibitem{ElkanN08}
Elkan, C., Noto, K.: Learning classifiers from only positive and unlabeled
  data. In: Li, Y., Liu, B., Sarawagi, S. (eds.) Proceedings of the 14th {ACM}
  {SIGKDD} International Conference on Knowledge Discovery and Data Mining, Las
  Vegas, Nevada, USA, August 24-27, 2008. pp. 213--220. {ACM} (2008).
  \doi{10.1145/1401890.1401920}, \url{https://doi.org/10.1145/1401890.1401920}

\bibitem{Friedman2001GBM}
Friedman, J.H.: Greedy function approximation: a gradient boosting machine.
  Annals of statistics pp. 1189--1232 (2001)

\bibitem{GranekHaber2015SDM}
Granek, J., Haber, E.: Data mining for real mining: A robust algorithm for
  prospectivity mapping with uncertainties. In: Proceedings of the 2015 {SIAM}
  International Conference on Data Mining. pp. 145--153. Society for Industrial
  and Applied Mathematics (SIAM) (2015). \doi{10.1137/1.9781611974010.17}

\bibitem{GranekHaber2016GeoscienceBC}
Granek, J., Haber, E.: Advanced geoscience targeting via focused machine
  learning applied to the {QUEST} project dataset, british columbia. In:
  Geoscience {BC} Summary of Activities 2015, pp. 117--126. Geoscience BC
  (2016), geoscience BC Report 2016-1

\bibitem{Ho2020DDPM}
Ho, J., Jain, A., Abbeel, P.: Denoising diffusion probabilistic models. In:
  Advances in Neural Information Processing Systems (NeurIPS) (2020).
  \doi{10.48550/arXiv.2006.11239}, \url{https://arxiv.org/abs/2006.11239}

\bibitem{HronskyKreuzer2019}
Hronsky, J.M.A., Kreuzer, O.P.: Applying spatial prospectivity mapping to
  exploration targeting: Fundamental practical issues and suggested solutions
  for the future. Ore Geology Reviews  \textbf{107},  647--653 (Apr 2019).
  \doi{10.1016/j.oregeorev.2019.03.016}

\bibitem{IyerNathSarawagi2014}
Iyer, A., Nath, S., Sarawagi, S.: Maximum mean discrepancy for class ratio
  estimation: Convergence bounds and kernel selection. In: International
  conference on machine learning. pp. 530--538. PMLR (2014)

\bibitem{Kirillov2020PointRend}
Kirillov, A., Wu, Y., He, K., Girshick, R.: Pointrend: Image segmentation as
  rendering. In: Proceedings of the IEEE/CVF Conference on Computer Vision and
  Pattern Recognition (CVPR). pp. 9799--9808 (2020).
  \doi{10.1109/CVPR42600.2020.00982},
  \url{https://openaccess.thecvf.com/content_CVPR_2020/html/Kirillov_PointRend_Image_Segmentation_As_Rendering_CVPR_2020_paper.html}

\bibitem{KiryoNPS17}
Kiryo, R., Niu, G., du~Plessis, M.C., Sugiyama, M.: Positive-unlabeled learning
  with non-negative risk estimator. In: Guyon, I., von Luxburg, U., Bengio, S.,
  Wallach, H.M., Fergus, R., Vishwanathan, S.V.N., Garnett, R. (eds.) Advances
  in Neural Information Processing Systems 30: Annual Conference on Neural
  Information Processing Systems 2017, December 4-9, 2017, Long Beach, CA,
  {USA}. pp. 1675--1685 (2017),
  \url{https://proceedings.neurips.cc/paper/2017/hash/7cce53cf90577442771720a370c3c723-Abstract.html}

\bibitem{Lipman2023FlowMatching}
Lipman, Y., Chen, R.T.Q., Ben-Hamu, H., Nickel, M., Le, M.: Flow matching for
  generative modeling. In: International Conference on Learning Representations
  (ICLR) (2023). \doi{10.48550/arXiv.2210.02747},
  \url{https://arxiv.org/abs/2210.02747}

\bibitem{LuEtAl2026Sparse3DGeoFM}
Lu, Z., Han, M., Guo, P., Bai, T., Su, J., Fang, F., Song, S.: Attention-guided
  flow-matching for sparse 3d geological generation. arXiv preprint
  arXiv:2604.09700  (2026)

\bibitem{McMillanTLE}
McMillan, M., Haber, E., Peters, B., Fohring, J.: Mineral prospectivity mapping
  using a vnet convolutional neural network. The Leading Edge  \textbf{40}(2),
  99--105 (02 2021). \doi{10.1190/tle40020099.1},
  \url{https://doi.org/10.1190/tle40020099.1}

\bibitem{PhillipsEtAl2009SampleBias}
Phillips, S.J., Dud{\'\i}k, M., Elith, J., Graham, C.H., Lehmann, A.,
  Leathwick, J., Ferrier, S.: Sample selection bias and presence-only
  distribution models: implications for background and pseudo-absence data.
  Ecological applications  \textbf{19}(1),  181--197 (2009)

\bibitem{duPlessisSugiyama2017}
Plessis, M.C., Niu, G., Sugiyama, M.: Class-prior estimation for learning from
  positive and unlabeled data. Mach. Learn.  \textbf{106}(4),  463--492 (Apr
  2017). \doi{10.1007/s10994-016-5604-6},
  \url{https://doi.org/10.1007/s10994-016-5604-6}

\bibitem{RamaswamyScottTewari2016}
Ramaswamy, H., Scott, C., Tewari, A.: Mixture proportion estimation via kernel
  embeddings of distributions. In: International conference on machine
  learning. pp. 2052--2060. PMLR (2016)

\bibitem{Schodde2025}
Schodde, R.: Mineral deposit exploration: Discovery trends, 1900-2023. SEG
  Discovery  \textbf{142},  19--35 (Jul 2025). \doi{10.5382/Geo-and-Mining-28}

\bibitem{SingerKouda1999}
Singer, D.A., Kouda, R.: Examining risk in mineral exploration. Natural
  Resources Research  \textbf{8}(2),  111--122 (1999).
  \doi{10.1023/A:1021838618750}

\bibitem{Song2021ScoreSDE}
Song, Y., Sohl-Dickstein, J., Kingma, D.P., Kumar, A., Ermon, S., Poole, B.:
  Score-based generative modeling through stochastic differential equations.
  In: International Conference on Learning Representations (ICLR) (2021).
  \doi{10.48550/arXiv.2011.13456}, \url{https://arxiv.org/abs/2011.13456}

\bibitem{Tong2023CFM}
Tong, A., Malkin, N., Huguet, G., Zhang, Y., Rector-Brooks, J., Fatras, K.,
  Wolf, G., Bengio, Y.: Conditional flow matching: Simulation-free dynamic
  optimal transport. arXiv preprint arXiv:2302.00482  (2023).
  \doi{10.48550/arXiv.2302.00482}, \url{https://arxiv.org/abs/2302.00482}

\bibitem{usgs_earthmri}
{U.S. Geological Survey}: Earth mapping resources initiative (earth mri).
  \url{https://www.usgs.gov/special-topics/earth-mri}, accessed: 2026-03-04

\bibitem{usgs_mrds}
{U.S. Geological Survey}: Mineral resources data system (mrds) (2026),
  \url{https://mrdata.usgs.gov/mrds/}, accessed: 2026-03-05

\bibitem{WartonShepherd2010PseudoAbsence}
Warton, D.I., Shepherd, L.C.: Poisson point process models solve the"
  pseudo-absence problem" for presence-only data in ecology. The Annals of
  Applied Statistics pp. 1383--1402 (2010)

\bibitem{Yang2019PointFlow}
Yang, G., Huang, X., Hao, Z., Liu, M.Y., Belongie, S., Hariharan, B.:
  Pointflow: 3d point cloud generation with continuous normalizing flows. In:
  Proceedings of the IEEE/CVF International Conference on Computer Vision
  (ICCV) (2019). \doi{10.1109/ICCV.2019.00464},
  \url{https://openaccess.thecvf.com/content_ICCV_2019/papers/Yang_PointFlow_3D_Point_Cloud_Generation_With_Continuous_Normalizing_Flows_ICCV_2019_paper.pdf}

\bibitem{Zuo2020}
Zuo, R.: Geodata science-based mineral prospectivity mapping: A review. Natural
  Resources Research  \textbf{29},  3415--3424 (2020).
  \doi{10.1007/s11053-020-09700-9}

\bibitem{ZuoWang2020}
Zuo, R., Wang, Z.: Effects of random negative training samples on mineral
  prospectivity mapping. Natural Resources Research  \textbf{29},  3443--3455
  (2020). \doi{10.1007/s11053-020-09668-6}

\end{thebibliography}

\clearpage
\title{Supplementary Material for\\ GeoCFM: Positive-Only Conditional Flow Matching for Mineral Occurrence Sampling}
\titlerunning{Supplementary Material: GeoCFM}
\author{Moshe Eliasof\inst{1}\thanks{Corresponding author.} \and Eldad Haber\inst{2}}
\authorrunning{M.~Eliasof and E.~Haber}
\institute{Faculty of Computer and Information Science, Ben-Gurion University of the Negev, Israel\\
\email{eliasof@bgu.ac.il} \and
Department of Earth, Ocean and Atmospheric Sciences,\\ University of British Columbia, Canada\\
\email{eldadhaber@gmail.com}}
\maketitle

\appendix

\section{Additional Experimental Details}
\label{sec:appendix}

This appendix provides additional details on the model, the benchmark construction, the real-data preprocessing pipeline, the baselines, and the evaluation protocol. We also include additional qualitative comparisons on held-out Earth MRI patches.

Throughout, occurrence locations are represented as point coordinates \((x,y)\), where \(x\) denotes the column index and \(y\) denotes the row index. During training, coordinates are normalized to \([0,1]^2\). For visualization and evaluation, they are mapped back to pixel coordinates.

\section{Implementation Details for GeoCFM}
\label{sec:appendix_method}

GeoCFM learns a conditional velocity field
\begin{equation}
v_\theta(x_t,t\mid d)\in\mathbb{R}^2
\end{equation}
that transports samples from a base Gaussian to the conditional distribution of occurrence locations given the input geo-image \(d\). Training follows the standard conditional flow-matching construction with linear interpolation
\begin{equation}
x_t = t x_1 + (1-t) z,
\qquad
z\sim\mathcal{N}(0,I_2),
\qquad
t\sim \mathrm{Unif}[0,1],
\end{equation}
and target velocity
\begin{equation}
v^\star = x_1 - z.
\end{equation}
The training objective is
\begin{equation}
\mathcal{L}_{\mathrm{CFM}}
=
\mathbb{E}
\left[
\left\|
v_\theta(x_t,t\mid d) - (x_1-z)
\right\|_2^2
\right].
\end{equation}

\subsection{Image-conditioned point transport}
The model combines an image encoder with a point-wise velocity predictor. A UNet first extracts a dense feature map
\begin{equation}
\Phi_\xi(d)\in\mathbb{R}^{D\times H\times W}.
\end{equation}
For each noised point \(x_t\), local features are then sampled from \(\Phi_\xi(d)\) by bilinear interpolation. If \(\phi_\xi(x_t;d)\in\mathbb{R}^D\) denotes the sampled feature vector and \(e(t)\in\mathbb{R}^{d_t}\) denotes a sinusoidal time embedding, the conditional velocity is given by
\begin{equation}
v_\theta(x_t,t\mid d)
=
h_\psi\!\left(
\left[
\phi_\xi(x_t;d),\,
x_t,\,
e(t)
\right]
\right),
\end{equation}
where \(h_\psi\) is a shared point-wise network.

This design combines multi-scale image context from the UNet with local information at the current point location.

\subsection{UNet backbone and velocity head}
The image encoder is a UNet built from residual convolutional blocks. Each block uses two \(3\times 3\) convolutions, GroupNorm, SiLU activations, and a residual skip connection. Downsampling is performed with \(2\times 2\) average pooling. In the decoder, features are upsampled bilinearly, projected with a \(1\times 1\) convolution, concatenated with the corresponding skip features, and passed through another residual block. A final \(1\times 1\) projection followed by SiLU produces the dense feature tensor used for point conditioning.

The time embedding has dimension \(64\). On Earth MRI, the velocity field is parameterized by a multilayer perceptron. On the synthetic benchmark, the final model uses a residual fully connected head: an input projection is followed by residual blocks with LayerNorm and SiLU nonlinearities, and then by a final LayerNorm and linear projection to \(\mathbb{R}^2\). This residual head is the version used in the final synthetic experiments.

We use the same UNet template on both benchmarks, with benchmark-specific widths. The exact choices are listed in Table~\ref{tab:appendix_geocfm_arch}.

\begin{table}[t]
\centering
\caption{GeoCFM architecture settings used in the two benchmarks.}
\label{tab:appendix_geocfm_arch}
\small
\setlength{\tabcolsep}{7pt}
\begin{tabular}{lcc}
\toprule
Setting & Synthetic & Earth MRI \\
\midrule
Input channels & 2 (\(d,t\)) & 8 geophysical layers \\
Base UNet width \(nf\) & 32 & 48 \\
UNet depth & 3 & 3 \\
Feature dimension \(D\) & 64 & 128 \\
Time-embedding dimension & 64 & 64 \\
Velocity head & residual network & MLP \\
Velocity hidden width & 256 & 256 \\
Velocity depth & 3 residual blocks & 3 hidden layers \\
\bottomrule
\end{tabular}
\end{table}

\subsection{Coordinate handling}
At training time, all points from a batch are concatenated into a single tensor, together with an index vector that records which image each point belongs to. Coordinates are normalized to \([0,1]^2\) before training. During sampling, the ODE is integrated in normalized coordinates and the final samples are mapped back to pixel space by multiplying by patch width and height.

\subsection{Input normalization}
Each input channel is standardized independently. Means and standard deviations are computed over non-zero pixels only, so zero-filled no-data regions do not affect the normalization. After standardization, zero-filled pixels remain zero. The same convention is used in both the synthetic and Earth MRI experiments.

\subsection{Optimization and sampling}
At test time, sampling is performed with explicit Euler integration,
\begin{equation}
x^{(k+1)} = x^{(k)} + \Delta t\, v_\theta(x^{(k)}, t_k \mid d),
\qquad
t_k = k\Delta t,
\qquad
\Delta t = 1/K.
\end{equation}
Final samples are clipped to the valid image domain before conversion back to pixel coordinates.

All models are trained with AdamW, cosine learning-rate decay, gradient clipping at norm \(1\), and GPU acceleration. The exact optimization settings are listed in Table~\ref{tab:appendix_geocfm_opt}.

\begin{table}[t]
\centering
\caption{GeoCFM optimization settings used in the two benchmarks.}
\label{tab:appendix_geocfm_opt}
\small
\setlength{\tabcolsep}{7pt}
\begin{tabular}{lcc}
\toprule
Setting & Synthetic & Earth MRI \\
\midrule
Batch size & 4 & 8 \\
Learning rate & \(3\times 10^{-4}\) & \(3\times 10^{-4}\) \\
Weight decay & \(10^{-4}\) & \(10^{-4}\) \\
Maximum iterations & 500{,}000 & 400{,}000 \\
Gradient clip & 1.0 & 1.0 \\
Validation interval & 250 iterations & 2{,}000 iterations \\
Visualization interval & 250 iterations & 10{,}000 iterations \\
Euler steps at test time & 50 & 50 \\
Samples per draw & 500 & 300 \\
\bottomrule
\end{tabular}
\end{table}

\section{Synthetic Magnetics-Geochemistry Benchmark}
\label{sec:appendix_synth}

The synthetic benchmark used in the final experiments is the second model problem, designed around RTP magnetics and observed geochemistry. Each example contains an observed magnetic image \(d\), an observed geochemistry proxy \(t\), a latent geochemical control \(s\), and a set of mineral occurrence coordinates.

The observed magnetic channel is built from a collection of elliptical intrusion-like bodies, yielding an RTP-style magnetic image with multiple candidate structures. In parallel, the generator produces an observed geochemistry channel \(t\) together with a latent geochemical field \(s\). The model is conditioned only on \((d,t)\); the latent field \(s\) is never shown to the network.

Mineralization is restricted to a subset of the candidate bodies through a latent activation mechanism. Thus, even when several intrusion-like bodies are visible in the observed magnetic field, only some are active, and within an active body the deposits localize along spatially concentrated arcs. This is what makes the mapping from the observed fields to the occurrence locations genuinely one-to-many.

The final synthetic setting used in the experiments has the following parameters:
\begin{itemize}
\item image size \(220\times 220\),
\item number of candidate bodies \(M=5\),
\item number of sampled deposits \(N=500\) per image,
\item two observed channels \((d,t)\),
\item a fixed held-out validation set generated with seeds \(800000,800001,\ldots\).
\end{itemize}

\subsection{Training and validation protocol}
Synthetic training examples are generated online throughout optimization, so the model is never trained on a fixed finite set of synthetic images. Validation is carried out on a fixed set of \(16\) held-out examples generated once at the start of training. The synthetic results in the main paper correspond to the final model based on the UNet image encoder and residual velocity head described above.

\section{Earth MRI Data, Preprocessing, and Spatial Split}
\label{sec:appendix_real}

\subsection{Raster construction}
The Earth MRI benchmark is built from DS-9 companion geophysical grids together with MRDS mineral occurrence records. All geophysical layers are rasterized onto a common longitude-latitude grid by binning each xyz file to cells of size \(0.02^\circ\times 0.02^\circ\). When multiple values fall into the same cell, they are averaged.

The implementation uses the following eight geophysical layers: \texttt{bouguer}, \texttt{isograv},
\texttt{magnetic}, \texttt{topo}, \texttt{topobaty}, \texttt{usak}, \texttt{usath}, and \texttt{usau}.
The final raster extent is determined automatically from the union of the DS-9 grids and covers the full raster footprint used in our experiments.

\subsection{Gap filling and no-data handling}
After binning to a regular longitude-latitude raster, many interior pixels remain empty because the original DS-9 grids are sampled on a regular Albers Equal-Area lattice but stored as longitude-latitude-value triples. These internal gaps are filled by iterative inverse-distance weighting, using radius \(5\) pixels, power \(2\), and \(5\) passes. Pixels that remain unsupported after these passes are left at the fill value \(0\). These zeros define the no-data region outside the valid survey footprint and are excluded from standardization and sampling.

\subsection{Occurrence alignment}
MRDS occurrence records are mapped to the same raster by converting longitude and latitude to pixel indices with flooring and clipping. The occurrence table is filtered by commodity using a case-insensitive match over the commodity fields, with the default filter \(\{\texttt{Gold},\texttt{Copper}\}\). Only occurrences inside the DS-9 raster bounding box are retained.

\subsection{Patch extraction}
Training and evaluation are performed on local patches extracted from the full raster. A patch is retained only if at least \(50\%\) of its pixels belong to the valid data footprint, defined as pixels where at least one channel is non-zero. The patch extraction settings are:
\begin{itemize}
\item patch size \(128\times 128\),
\item stride \(16\),
\item minimum valid-footprint fraction \(0.5\).
\end{itemize}
For each retained patch, the occurrence annotations are converted into a patch-local deposit-count image. When multiple deposits fall in the same raster cell, the count at that pixel is greater than one. During flow-matching training and evaluation, these counts are converted back into repeated point coordinates.

\subsection{Spatially disjoint train-test split}
To reduce spatial leakage from geographic autocorrelation, train and test patches are assigned by large spatial tiles rather than independently. The full raster is partitioned into \(384\times 384\) tiles, and \(20\%\) of the tiles are assigned to the held-out split using random seed \(42\).

A buffer-zone rule is then applied: a patch is kept in a split only if all tiles touched by the full spatial extent of that patch belong to that split. Patches that straddle train-test boundaries are discarded. This yields a stricter protocol than assigning patches by their top-left corners alone.

\subsection{Additional qualitative comparisons on Earth MRI}
\begin{figure*}[t]
\centering
\includegraphics[width=\linewidth]{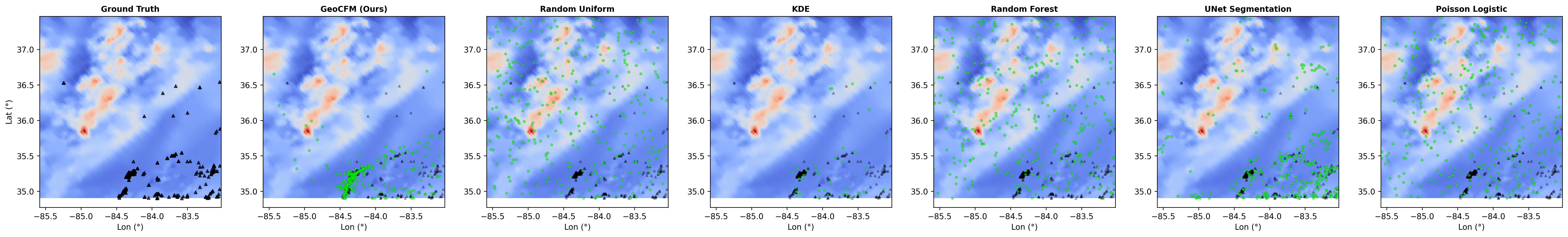}\vspace{2mm}

\includegraphics[width=\linewidth]{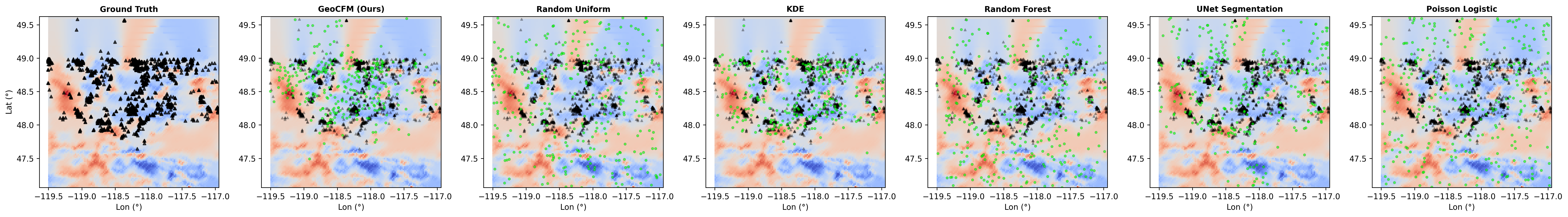}\vspace{2mm}

\includegraphics[width=\linewidth]{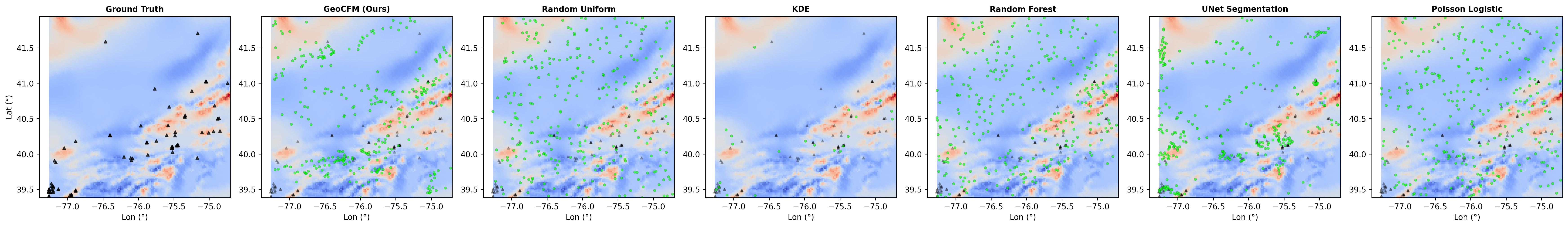}
\caption{\textbf{Additional qualitative comparisons on held-out Earth MRI patches.} Each row shows one held-out patch. From left to right, the columns correspond to ground truth, GeoCFM, Random Uniform, KDE, Random Forest, UNet Segmentation, and Poisson Logistic. Black triangles denote observed mineral occurrences, and green points denote sampled occurrences. GeoCFM produces samples that are more spatially concentrated around the observed structures than the unconditional or score-map baselines, while still expressing uncertainty through repeated draws.}
\label{fig:appendix_ds9_qual}
\end{figure*}

Figure~\ref{fig:appendix_ds9_qual} shows additional held-out Earth MRI patches together with samples drawn from GeoCFM and from all baselines. In each row, the first column shows the ground-truth occurrences, and the remaining columns show the corresponding method outputs on the same held-out patch.

\section{Baseline Details}
\label{sec:appendix_baselines}

All baselines are evaluated in the same point-sampling framework as GeoCFM. Each method defines a spatial distribution over pixels, from which a point set is sampled with replacement and perturbed by sub-pixel jitter. For cross-method comparison, each method generates exactly as many points as there are observed deposits in the current image or patch.

\subsection{Uniform}
The uniform baseline samples points uniformly over the admissible spatial domain. On the synthetic benchmark, this is the full image canvas. On Earth MRI, sampling is restricted to the valid survey footprint of the patch, defined by pixels for which at least one channel is non-zero.

\subsection{Global KDE}
The KDE baseline is unconditional. It fits a two-dimensional Gaussian kernel density estimate to the training deposit coordinates and ignores the input image. On Earth MRI, the estimator is fit in global raster coordinates, after removing duplicates introduced by overlapping training patches. The bandwidth parameter is set through
\begin{equation}
\texttt{bw\_method} = \frac{5.0}{\mathrm{std}(\mathcal{P}_{\mathrm{train}})},
\end{equation}
where \(\mathcal{P}_{\mathrm{train}}\) denotes the set of training deposit coordinates.

At test time, KDE samples are drawn in global coordinates and only those falling inside the queried patch are retained. If too few accepted samples are obtained, the remaining points are filled with uniform samples inside the patch.

\subsection{Poisson-LR}
Poisson-LR treats prospectivity mapping as pixel-wise intensity estimation using logistic regression on local features. On the synthetic benchmark, the feature vector at each pixel is the two-channel input \([d,t]\). On Earth MRI, it is the full multi-channel geophysical vector at that pixel.

Positive pixels are those containing at least one observed occurrence. Unlabeled pixels are subsampled and treated as pseudo-negatives. On Earth MRI, the number of pseudo-negative pixels per training patch is capped at
\begin{equation}
\min\!\left(|\mathcal{N}|,\max(3N_+,50)\right),
\end{equation}
where \(N_+\) is the number of positive pixels and \(|\mathcal{N}|\) is the number of available unlabeled pixels.

The classifier uses the \texttt{lbfgs} solver, balanced class weights, \(C=1.0\), and a maximum of \(1000\) iterations. The predicted positive probabilities are treated as an unnormalized intensity map and normalized over the valid pixels before sampling.

\subsection{Random Forest}
The Random Forest baseline is also trained as a pixel-wise classifier. On the synthetic benchmark, the feature vector at each pixel is the two-channel input \([d,t]\). On Earth MRI, it is the full multi-channel feature vector. Positive pixels correspond to deposit pixels, and unlabeled pixels are subsampled as pseudo-negatives.

On Earth MRI, the number of pseudo-negative pixels per training patch is capped at
\begin{equation}
\min\!\left(|\mathcal{N}|,\max(3N_+,100)\right).
\end{equation}
The classifier uses \(200\) trees, maximum depth \(15\), balanced class weights, and random seed \(42\). At test time, predicted positive probabilities are normalized over the admissible domain and sampled to produce point locations.

\subsection{UNet-Seg}
The segmentation baseline predicts a dense per-pixel occurrence score map and then samples points proportionally to the predicted probabilities. To keep the comparison architecture-aware, it uses the same UNet backbone family as GeoCFM, with residual convolutional blocks, GroupNorm, SiLU activations, average-pooling downsampling, bilinear upsampling, and skip connections. The main difference is that the point-conditioned transport head is replaced by a single-channel segmentation head.

On Earth MRI, the segmentation model uses \(nf=48\), depth \(3\), AdamW with learning rate \(3\times 10^{-4}\), weight decay \(10^{-4}\), binary cross-entropy with logits, positive-class weight \(50\), cosine learning-rate decay, gradient clipping at norm \(1\), batch size \(8\), and \(50{,}000\) training iterations. Model selection is based on the best validation loss measured every \(2{,}000\) iterations.

\subsection{GBDT}
The gradient-boosted decision tree baseline is trained as a pixel-wise classifier on the same per-pixel
features as Random Forest, using positives and subsampled pseudo-negatives. It uses gradient-boosted
trees with the same point-sampling protocol as the other score-map methods: predicted positive
probabilities are normalized over the admissible domain and sampled to produce point locations.

\subsection{OCSVM}
The one-class SVM baseline is a positive-only model that does not use pseudo-negatives. It is fit on
the per-pixel feature vectors at observed occurrence locations to estimate the support of the positive
distribution with an RBF kernel. At test time, the decision-function scores over all admissible pixels
are shifted to be non-negative, normalized into a spatial distribution, and sampled in the same way as
the score-map baselines.

\subsection{Retrieval-KDE}
The retrieval-based KDE baseline is also positive-only. For each query patch, it retrieves training
occurrence locations and places a Gaussian kernel density estimate over the retrieved points, which is
then normalized over the admissible domain and sampled. Unlike the global KDE, the density is induced
by retrieval rather than fit once over all training occurrences, but it still does not condition on the
input geo-image through a learned encoder.

\subsection{From score maps to point sets}
For Poisson-LR, Random Forest, GBDT, UNet-Seg, OCSVM, and Retrieval-KDE, the predicted score map is converted into a probability distribution by clipping scores below by a small \(\varepsilon\), masking invalid pixels when appropriate, and normalizing over the spatial domain. Point samples are then drawn with replacement from the resulting multinomial distribution and perturbed by independent uniform jitter in \([-0.5,0.5]\) pixels in each coordinate.

\section{Evaluation Protocol}
\label{sec:appendix_eval}

\subsection{Chamfer distance}
All methods are evaluated using the symmetric Chamfer distance between generated and observed point sets:
\begin{equation}
\mathrm{CD}(A,B)
=
\frac{1}{|A|}
\sum_{a\in A}
\min_{b\in B}\|a-b\|_2
+
\frac{1}{|B|}
\sum_{b\in B}
\min_{a\in A}\|b-a\|_2.
\end{equation}
Distances are computed in pixel coordinates. Pairwise distances are accumulated in float64 for numerical robustness.

\subsection{Additional metrics}
Beyond Chamfer distance, we report four complementary metrics, all computed between the generated and
observed point sets (or densities derived from them). The \emph{Sinkhorn} divergence is an
entropy-regularized optimal-transport distance between the two point sets, capturing distribution
matching rather than nearest-neighbor agreement. \emph{F@5} is the F-score obtained by matching generated
and observed points within a $5$-pixel tolerance via bipartite assignment. The \emph{KDE negative
log-likelihood} (NLL) fits a Gaussian KDE to the generated samples and scores the observed occurrences
under it, so lower values indicate that observed deposits lie in high-density regions of the predicted
distribution. The \emph{top-5\% hit rate} is the fraction of observed occurrences that fall within the
top $5\%$ most prospective pixels of the (normalized) predicted density. Each metric is averaged over
$20$ stochastic draws per input and reported as mean $\pm$ standard deviation across held-out examples.

\subsection{Point-count protocol}
For the main cross-method comparison, each method is sampled to produce exactly as many points as there are observed deposits in the current image or patch. This keeps the comparison focused on spatial quality rather than point-count mismatch.

\subsection{Repeated draws}
Qualitative results are reported as repeated independent draws for the same input. For the synthetic benchmark, each qualitative draw uses \(500\) sampled points. For Earth MRI, each qualitative draw uses \(300\) sampled points.

\subsection{Train--test generalization gap}
\Cref{tab:traintest_gap} reports mean Chamfer distance on the train and test splits of both benchmarks.
GeoCFM exhibits a small train--test gap (synthetic $8.81\!\to\!9.37$, Earth MRI $11.89\!\to\!12.30$),
comparable in relative terms to the baselines despite its much lower absolute error, supporting the
claim that its advantage reflects genuine conditional generalization rather than overfitting to training
tiles (Q3).

\begin{table}[t]
\centering
\caption{Train vs.\ test Chamfer distance (pixels, lower is better) on both benchmarks. The Earth MRI
split is spatially disjoint (tile-based).}
\label{tab:traintest_gap}
\footnotesize
\setlength{\tabcolsep}{6pt}
\begin{tabular}{lcccc}
\toprule
& \multicolumn{2}{c}{Synthetic} & \multicolumn{2}{c}{Earth MRI} \\
\cmidrule(lr){2-3}\cmidrule(lr){4-5}
Method & Train CD $\downarrow$ & Test CD $\downarrow$ & Train CD $\downarrow$ & Test CD $\downarrow$ \\
\midrule
Uniform sampling & 49.30 & 49.61 & 63.99 & 64.25 \\
Global KDE & 37.12 & 41.16 & 57.10 & 66.35 \\
Poisson-LR & 38.75 & 39.93 & 62.80 & 64.21 \\
Random Forest & 32.08 & 37.09 & 49.12 & 54.37 \\
UNet-Seg & 26.11 & 29.46 & 41.30 & 46.81 \\
\midrule
GeoCFM (Ours) & 8.81 & 9.37 & 11.89 & 12.30 \\
\bottomrule
\end{tabular}
\end{table}

\section{Additional Ablation Results}
\label{sec:appendix_ablation}
This section provides the full tables for the analyses summarized in Section~5.3 of the main paper:
the positive-subset stability study (\Cref{tab:subset}), the inference-cost / Euler-step ablation
(\Cref{tab:runtime}), and the sampling-budget study (\Cref{tab:n}).

\begin{table}[t]
\centering
\caption{Positive-subset stability. GeoCFM is retrained with different fractions of the available
positives over three seeds; we report mean $\pm$ std across seeds.}
\label{tab:subset}
\footnotesize
\setlength{\tabcolsep}{6pt}
\resizebox{\linewidth}{!}{%
\begin{tabular}{llcccc}
\toprule
Data & Pos.\ frac. & CD $\downarrow$ & Sink. $\downarrow$ & NLL $\downarrow$ & F@5 $\uparrow$ \\
\midrule
Synthetic & $100\%$ & $9.38\pm0.09$ & $0.0447\pm0.0006$ & $8.96\pm0.07$ & $0.633\pm0.005$ \\
Synthetic & $75\%$  & $9.62\pm0.10$ & $0.0457\pm0.0006$ & $8.87\pm0.06$ & $0.636\pm0.005$ \\
Synthetic & $50\%$  & $9.60\pm0.14$ & $0.0460\pm0.0010$ & $9.01\pm0.06$ & $0.627\pm0.005$ \\
\midrule
Earth MRI & $100\%$ & $12.35\pm0.21$ & $0.085\pm0.002$ & $10.76\pm0.15$ & $0.151\pm0.006$ \\
Earth MRI & $75\%$  & $11.91\pm0.17$ & $0.076\pm0.002$ & $10.63\pm0.16$ & $0.189\pm0.007$ \\
Earth MRI & $50\%$  & $12.46\pm0.19$ & $0.084\pm0.002$ & $10.39\pm0.11$ & $0.159\pm0.007$ \\
\bottomrule
\end{tabular}%
}
\end{table}

\begin{table}[t]
\centering
\caption{Inference cost vs.\ number of Euler steps $K$, compared with the single-pass UNet-Seg baseline.
Runtime is milliseconds per patch.}
\label{tab:runtime}
\footnotesize
\setlength{\tabcolsep}{6pt}
\begin{tabular}{llccc}
\toprule
Data & Method & ms/patch $\downarrow$ & CD $\downarrow$ & F@5 $\uparrow$ \\
\midrule
Synthetic & UNet-Seg & 13.92 & 29.46 & 0.615 \\
Synthetic & GeoCFM $K=5$  & 19.79 & 9.30 & 0.580 \\
Synthetic & GeoCFM $K=10$ & 21.98 & 9.39 & 0.610 \\
Synthetic & GeoCFM $K=50$ & 38.47 & 9.37 & 0.634 \\
\midrule
Earth MRI & UNet-Seg & 12.50 & 46.81 & 0.085 \\
Earth MRI & GeoCFM $K=5$  & 16.95 & 12.00 & 0.140 \\
Earth MRI & GeoCFM $K=10$ & 21.68 & 11.85 & 0.155 \\
Earth MRI & GeoCFM $K=50$ & 58.99 & 12.30 & 0.151 \\
\bottomrule
\end{tabular}
\end{table}

\begin{table}[t]
\centering
\caption{Effect of the Monte Carlo sampling budget $N$ on density quality (KDE-NLL, lower is better) and
density stability (higher is better).}
\label{tab:n}
\footnotesize
\setlength{\tabcolsep}{6pt}
\begin{tabular}{lccccc}
\toprule
Budget $N$ & 100 & 250 & 500 & 1000 & 2000 \\
\midrule
Synthetic NLL $\downarrow$ & 11.79 & 10.96 & 10.79 & 10.66 & 10.56 \\
Synthetic stab. $\uparrow$ & 0.894 & 0.946 & 0.970 & 0.982 & 0.988 \\
Earth MRI NLL $\downarrow$ & 10.20 & 9.32 & 9.24 & 9.02 & 8.99 \\
Earth MRI stab. $\uparrow$ & 0.876 & 0.962 & 0.971 & 0.984 & 0.991 \\
\bottomrule
\end{tabular}
\end{table}

\end{document}